\documentclass{article}

\PassOptionsToPackage{numbers,sort&compress}{natbib}
\usepackage{graphicx}
\usepackage{multirow}
\usepackage[ruled,vlined]{algorithm2e}
\SetKwInOut{Input}{Input}
\SetKwInOut{Output}{Output}
\usepackage{amsmath}
\usepackage{placeins}
\usepackage[table]{xcolor}
\usepackage{xcolor}
\newcommand{\epimark}{\textcolor{red}{\ensuremath{^{\dagger}}}}

 \usepackage[preprint]{format}

\usepackage[utf8]{inputenc} 
\usepackage[T1]{fontenc}    
\usepackage{hyperref}       
\usepackage{url}            
\usepackage{booktabs}       
\usepackage{amsfonts}       
\usepackage{nicefrac}       
\usepackage{microtype}      
\usepackage{xcolor}         

\title{Frequency-Structured Field Learning for Light-Field Disparity Estimation}

\author{%
  Sara Monji-Azad\thanks{Corresponding author: \texttt{sara.monjiazad@medma.uni-heidelberg.de}} \\
  Mannheim Institute for Intelligent Systems in Medicine (MIISM)\\
  Medical Faculty Mannheim, Heidelberg University, Mannheim, Germany\\
  \And
  Yulin Liu \\
  Mannheim Institute for Intelligent Systems in Medicine (MIISM)\\
  Medical Faculty Mannheim, Heidelberg University, Mannheim, Germany\\
  \AND
  J\"urgen Hesser \\
  Mannheim Institute for Intelligent Systems in Medicine (MIISM), \\Medical Faculty Mannheim, Heidelberg University, Mannheim, Germany\\
  Interdisciplinary Center for Scientific Computing (IWR), Heidelberg University\\
  Central Institute for Computer Engineering (ZITI), Heidelberg University\\
  CZS Heidelberg Center for Model-Based AI, Heidelberg University\\
  Germany\\
}

\begin{document}

\maketitle

\begin{abstract}
Light-field disparity estimation requires both global consistency over smooth or textureless regions and local precision near occlusion boundaries, thin structures, and abrupt depth transitions. Existing methods address these requirements mainly through local EPI matching, cost-volume or focal-stack construction, view aggregation, or direct convolutional regression. While effective, these designs often rely on local matching windows, discrete disparity hypotheses, memory-intensive volumes, or attention-based aggregation. This motivates an alternative field-level formulation: predicting disparity from globally and locally updated EPI-derived latent features rather than from an explicitly constructed disparity volume. We introduce FreqLF, an EPI-guided Fourier-local framework based on this principle. FreqLF encodes angular parallax cues from horizontal and vertical EPI stacks together with central-view appearance features, projects them to a latent feature field, and updates this field through stacked hybrid Fourier-local layers. Fourier-domain low-mode updates provide global feature interaction, while local convolutional updates preserve spatial variations needed for fine disparity detail. Disparity is decoded from the updated latent field using a coordinate-conditioned Gaussian-mixture decoder, with the mixture mean used as the final prediction. Experiments on the HCI 4D Light Field Benchmark show that FreqLF approaches the accuracy of strong supervised baselines while avoiding explicit cost-volume construction in the base model. Ablations indicate complementary contributions of the Fourier and local branches, and scaling experiments show practical behavior across spatial resolutions. These results suggest that Fourier-local latent field learning offers a competitive alternative design principle for light-field disparity estimation, trading explicit disparity-volume construction for global-local feature updates over EPI-derived latent fields. (The code will be published soon.)

\end{abstract}

\section{Introduction}
Light-field cameras capture both spatial appearance and angular variation of incoming rays, enabling dense multi-view reasoning from a single acquisition \citep{levoy2023light,gortler2023lumigraph,ng2005light}. Under the standard two-plane parameterization, a light field is represented over spatial and angular coordinates, and epipolar plane images (EPIs) reveal a direct geometric relationship between line slope and disparity \citep{bolles1987epipolar,adelson1992single,shin2018epinet}. This makes light fields attractive for depth estimation, 3D reconstruction, computational photography, robotic perception, refocusing, and view synthesis \citep{honauer2016dataset,shin2018epinet,tsai2020attention,wang2022occlusion}.

Despite this rich angular structure, robust light-field disparity estimation remains challenging. Smooth surfaces, textureless regions, and large planar areas require globally consistent disparity estimates, while occlusion boundaries, thin structures, repetitive patterns, and abrupt foreground-background transitions require local precision \citep{zhang2016robust,shin2018epinet,wang2022occlusion}. Textureless regions provide weak local evidence, and occluded regions produce inconsistent observations across viewpoints \citep{tsai2020attention,li2023opal,gao2025epipolar}. These challenges indicate that light-field disparity is not simply a pixel-wise regression problem. The target disparity map is a structured spatial field that should combine long-range consistency with local spatial detail.

Current light-field disparity formulations largely focus on where to match across views or how to aggregate angular evidence. Local EPI and convolutional methods exploit short-range angular patterns \citep{zhang2016robust,shin2018epinet,gao2022epi}; cost-volume and focal-stack methods explicitly evaluate candidate disparities and regularize matching evidence \citep{wang2022occlusion,chao2023learning,liu2022cascade}; attention-based methods select or aggregate reliable angular evidence \citep{tsai2020attention,li2023opal,gao2022epi}; and recent implicit neural field approaches represent disparity continuously but often rely on iterative optimization \citep{shi2025iterative}. These approaches have substantially advanced light-field disparity estimation. However, they often rely on local matching windows, discrete disparity hypotheses, memory-intensive volumes, attention-based aggregation, or iterative optimization. This motivates an alternative field-level formulation: predicting disparity from globally and locally updated EPI-derived latent features rather than from an explicitly constructed disparity volume.

We introduce FreqLF, an EPI-guided Fourier-local framework for light-field disparity estimation. FreqLF first extracts angular parallax cues from horizontal and vertical EPI stacks, together with central-view appearance features. The resulting spatial feature field is projected to a latent feature field and updated by stacked hybrid Fourier-local layers. Each layer updates the same latent representation through two complementary branches: a Fourier-domain branch that applies learned weights to retained low-frequency Fourier modes for global feature interaction, and a local convolutional branch that applies spatial $3 \times 3$ updates to preserve local spatial variations needed for fine disparity detail. A coordinate-conditioned Gaussian-mixture decoder then predicts disparity from the updated latent field, with the mixture mean used as the final disparity estimate. FreqLF therefore shifts the main computation from explicit disparity-volume construction to latent field updating. The Fourier and convolutional branches operate on a shared EPI-derived latent feature field: the Fourier branch introduces low-mode global feature interaction, while the local branch provides spatially localized updates. Their outputs are combined at each layer, producing a representation that carries both long-range field context and local spatial detail before disparity decoding. In its base form, this allows FreqLF to estimate disparity without explicitly enumerating candidate disparities in a cost volume. We evaluate FreqLF on the HCI 4D Light Field Benchmark \citep{honauer2016dataset}. The results show that FreqLF approaches the accuracy of strong supervised baselines while avoiding explicit cost-volume construction in the base model. Beyond average accuracy, we analyze the proposed formulation through branch ablations, Fourier-local complementarity, scaling behavior, and region-specific diagnostics. These experiments suggest that Fourier-local latent field learning is a practical alternative design principle for light-field disparity estimation, especially when global field interaction is desired without explicit disparity-volume construction. In summary, our contributions are:
\begin{itemize}
    \item We proposed a cost-volume-free, field-level formulation for light-field disparity estimation, where disparity is decoded from EPI-derived latent feature fields updated by complementary global Fourier-domain and local convolutional operations.

    \item We introduce FreqLF, an EPI-guided Fourier-local architecture that encodes horizontal and vertical angular parallax cues, projects the fused spatial features to a latent field, and updates this field through stacked hybrid Fourier-local layers before disparity decoding.

    \item We demonstrate that the base FreqLF model approaches the accuracy of strong supervised light-field disparity baselines without explicit cost-volume construction, showing that discrete disparity-volume enumeration can be traded for global-local feature updates over EPI-derived latent fields.

    \item We analyze the proposed design through branch ablations, Fourier-local complementarity, scaling experiments, and region-specific diagnostics, identifying both the benefits of the Fourier-local mechanism and the remaining limitations of the current model.
\end{itemize}
\section{Related Work}

\paragraph{Geometry-driven light-field disparity.}
Light fields provide a dense sampling of scene radiance over spatial and angular dimensions \citep{levoy2023light,gortler2023lumigraph,ng2005light}. Their epipolar plane images (EPIs) expose a direct relationship between angular line slope and disparity \citep{bolles1987epipolar,adelson1992single}. Classical light-field depth methods exploit this structure through EPI orientation analysis, structure tensors, variational regularization, defocus-correspondence cues, and geometric consistency \citep{wanner2012globally,wanner2013variational,tao2013depth,zhang2016robust}. These methods are interpretable and closely tied to light-field geometry, but their local or handcrafted assumptions make them less robust under occlusion, repeated textures, and thin structures. FreqLF retains the geometric advantage of EPI features, but uses them to condition a learned frequency-structured disparity field rather than to perform local slope selection.

\paragraph{Learned spatial-angular representations.}
Deep learning methods replace handcrafted EPI filters with learned spatial-angular representations. \citet{shin2018epinet} introduced EPINET, a fully convolutional network that processes EPI stacks from multiple angular directions and established deep EPI processing as a strong light-field disparity baseline. Later methods improve efficiency or representation capacity through lightweight EPINET variants, directional EPI modeling, and spatial-angular disentanglement \citep{hassan2022light,gao2022epi,wang2022disentangling}. These methods learn useful angular features, but remain primarily convolutional or feature-disentanglement pipelines. In contrast, FreqLF uses EPI-derived features as input to a spectral-local field operator, explicitly separating smooth global structure from high-frequency disparity discontinuities.

\paragraph{Cost-volume and occlusion-aware matching.}
Cost-volume methods estimate disparity by constructing matching evidence over candidate disparities and regularizing the resulting volume. This provides strong multi-view matching, but increases memory usage and ties prediction to discrete disparity hypotheses. \citet{wang2022occlusion} proposed OACC-Net, an occlusion-aware cost constructor for light-field depth estimation. \citet{chao2023learning} model sub-pixel disparity distributions using focal-stack and cost-volume information. Other recent methods refine cost construction through cascade estimation or EPI-guided cost volumes \citep{liu2022cascade,wang2023epi}. These methods remain centered on constructing and regularizing matching costs. In contrast, the base version of FreqLF avoids explicit cost-volume construction and replaces discrete disparity enumeration with field-level spectral-local reasoning over EPI-derived latent features.

\paragraph{Attention, view aggregation, and self-supervision.}
Attention-based methods improve robustness by selecting or aggregating reliable angular evidence. \citet{tsai2020attention} introduced attention-based view selection for light-field disparity estimation, while later methods combine attention with EPI constraints or multi-scale aggregation \citep{gao2022epi,chen2021attention,gao2025epipolar}. Self-supervised and unsupervised approaches reduce dependence on ground-truth disparity by using photometric consistency, occlusion-aware losses, or epipolar consistency \citep{li2023opal,gao2025epipolar}. These methods focus on view reliability, angular aggregation, or supervision design. FreqLF is complementary: it does not primarily ask which view to trust, but how the output disparity field should be represented and processed once angular evidence has been encoded.

\paragraph{Continuous fields and neural operators.}
Implicit neural representations model continuous signals and query outputs at arbitrary coordinates \citep{mildenhall2021nerf,sitzmann2020implicit}. Recent light-field disparity work has explored continuous neural disparity fields, often through iterative or test-time optimization \citep{shi2025iterative}. Neural operators instead learn mappings between function spaces, with Fourier Neural Operators providing global receptive fields through frequency-domain convolution \citep{li2020fourier,kovachki2023neural}. FreqLF takes inspiration from this operator view, but adapts it to light-field disparity through an EPI-guided hybrid design: global geometry is modeled by a Fourier branch, while boundary detail is preserved by a local convolutional branch. Thus, our goal is not merely continuous querying or applying a Fourier-local latent-field backbone, but frequency-structured disparity field learning.

\section{Proposed Method}
\label{sec:method}

\begin{figure}[t]
  \centering
  \includegraphics[width=13cm]{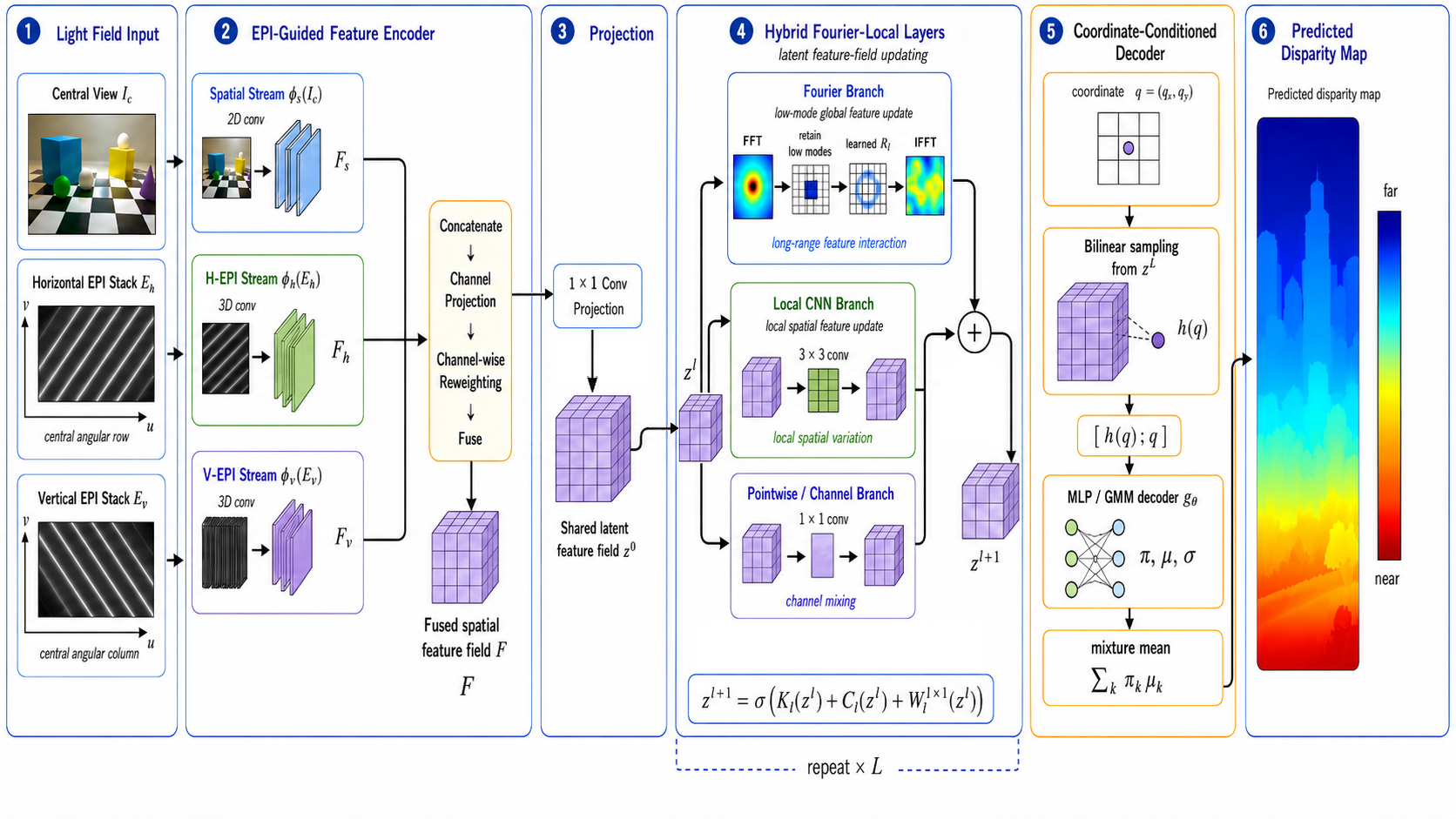}
    \caption{Overview of FreqLF. The central view and horizontal/vertical EPI stacks are encoded into a fused spatial feature field and projected to a latent feature field. Stacked hybrid Fourier-local layers update this shared field through Fourier-domain low-mode global feature interactions, local convolutional updates, and pointwise channel mixing; the final latent field is bilinearly sampled at image coordinates and decoded by a Gaussian-mixture MLP, whose mixture mean gives the disparity.}
\label{fig:freqlf_overview}
\end{figure}

\paragraph{Problem formulation.}
We consider a densely sampled light field
\begin{equation}
    \mathcal{L} \in \mathbb{R}^{A_v \times A_h \times H \times W \times 3},
\end{equation}
where $A_v$ and $A_h$ denote the number of angular views along the vertical and horizontal axes, and $H \times W$ is the spatial resolution of each sub-aperture image. The central view is denoted by $I_c \in \mathbb{R}^{3 \times H \times W}$. The goal is to estimate the central-view disparity field
\begin{equation}
    d: \Omega \rightarrow \mathbb{R}, 
    \qquad 
    \Omega = [-1,1]^2,
\end{equation}
which maps each normalized image coordinate $\mathbf{q}=(q_x,q_y)\in\Omega$ to a scalar disparity value. A common formulation is to infer $d(\mathbf{q})$ by local angular matching, direct regression from learned image features, or selection from a discrete set of candidate disparities. We instead view disparity estimation as prediction from a structured spatial field representation. The target disparity map should remain globally consistent over smooth surfaces, textureless regions, and large planar areas, while preserving sharp local changes at occlusion boundaries, thin structures, and foreground-background discontinuities. This creates a modeling tension. Local operators preserve spatial detail but have limited global context. Fourier-domain updates provide efficient full-field interaction through retained low-frequency modes, but low-mode truncation alone may lose fine spatial variations. FreqLF addresses this tension at the latent feature level: a Fourier-domain branch provides global feature interaction, while a local convolutional branch provides spatially localized feature updates before disparity decoding.

\paragraph{Overview of FreqLF.}
The central component of FreqLF is a stack of hybrid Fourier-local layers applied to an EPI-derived latent feature field. The EPI encoder provides angular evidence, the Fourier-local layers update the latent field globally and locally, and the decoder predicts disparity from the updated representation. FreqLF follows a three-stage design, illustrated in Fig.~\ref{fig:freqlf_overview}. First, an EPI-guided encoder extracts angular parallax cues from horizontal and vertical EPI stacks, together with appearance features from the central view. Second, the fused spatial feature field is projected to a latent feature field and updated by stacked hybrid Fourier-local layers. Third, a coordinate-conditioned decoder bilinearly samples the updated latent field and predicts disparity values.

We construct horizontal and vertical EPI stacks through the central angular row and central angular column:
\begin{equation}
    E_h = \mathcal{L}[u_c, :, :, :, :] \in \mathbb{R}^{A_h \times 3 \times H \times W},
    \qquad
    E_v = \mathcal{L}[:, v_c, :, :, :] \in \mathbb{R}^{A_v \times 3 \times H \times W}.
\end{equation}
Here, $u_c$ and $v_c$ denote the central vertical and horizontal angular indices. Each stack contains sub-aperture views along one central angular line. Classical 2D EPI slices are obtained by fixing a spatial row or column within these stacks. Given the central view $I_c$, horizontal EPI stack $E_h$, vertical EPI stack $E_v$, and query coordinates $\mathcal{Q}=\{\mathbf{q}_i\}_{i=1}^{N_q}$, FreqLF predicts
\begin{equation}
    \hat{d}(\mathbf{q}_i) = f_{\theta}(I_c, E_h, E_v, \mathbf{q}_i).
\end{equation}
In its base form, FreqLF avoids explicit cost-volume construction. It does not enumerate candidate disparities or build a multi-view matching volume. Instead, it encodes angular geometry into a spatial latent field and performs Fourier-local feature updating before disparity decoding.

\paragraph{EPI-guided feature encoding.}
The encoder maps the light-field input to a spatial feature field
\begin{equation}
    \mathbf{F} \in \mathbb{R}^{C_F \times H \times W}.
\end{equation}
It combines three complementary sources of evidence: central-view appearance, horizontal EPI geometry, and vertical EPI geometry. The central view is processed by a spatial stream
\begin{equation}
    \mathbf{F}_s = \phi_s(I_c),
    \qquad
    \mathbf{F}_s \in \mathbb{R}^{C_s \times H \times W},
\end{equation}
which extracts local texture, edge, and appearance features. The horizontal and vertical EPI stacks are processed by angular streams
\begin{equation}
    \mathbf{F}_h = \phi_h(E_h),
    \qquad
    \mathbf{F}_v = \phi_v(E_v),
    \qquad
    \mathbf{F}_h \in \mathbb{R}^{C_h \times H \times W},
    \qquad
    \mathbf{F}_v \in \mathbb{R}^{C_v \times H \times W}.
\end{equation}
The angular streams use 3D convolutions to aggregate information along the angular dimension while preserving spatial resolution. These streams encode EPI line-slope patterns, which provide direct geometric cues for disparity. The three feature maps are concatenated and fused:
\begin{equation}
    \mathbf{F}_{cat} = [\mathbf{F}_s;\mathbf{F}_h;\mathbf{F}_v],
    \qquad
    \mathbf{F} = \phi_f(\mathbf{F}_{cat}),
    \qquad
    \mathbf{F} \in \mathbb{R}^{C_F \times H \times W}.
\end{equation}
The fusion module uses a channel projection followed by channel-wise reweighting. This allows the model to adaptively balance appearance information from the central view and angular-geometric information from the EPI streams before applying Fourier-local field processing.

\paragraph{Hybrid Fourier-local layers.}
The encoder output is first projected to the latent feature dimension by a $1\times1$ convolution:
\begin{equation}
    \mathbf{z}^{0} = W_{\mathrm{proj}}(\mathbf{F}),
    \qquad
    \mathbf{z}^{0} \in \mathbb{R}^{C_z \times H \times W}.
\end{equation}
FreqLF then applies $L$ hybrid Fourier-local layers. Each layer updates the same latent feature field through a Fourier-domain branch, a local convolutional branch, and a pointwise residual projection:
\begin{equation}
    \mathbf{z}^{\ell+1}
    =
    \sigma\left(
        \mathcal{K}_{\ell}(\mathbf{z}^{\ell})
        +
        \mathcal{C}_{\ell}(\mathbf{z}^{\ell})
        +
        W_{\ell}^{1\times1}(\mathbf{z}^{\ell})
    \right),
    \qquad
    \ell=0,\ldots,L-1,
\end{equation}
where $\sigma$ is a nonlinear activation, $\mathcal{K}_{\ell}$ is the Fourier-domain low-mode update, $\mathcal{C}_{\ell}$ is the local convolutional update, and $W_{\ell}^{1\times1}$ is a pointwise channel-mixing projection. The Fourier branch provides a learned global update to the latent feature field. It applies a 2D Fourier transform over the spatial dimensions, learns complex-valued weights on a truncated set of low-frequency modes, and maps the result back to the spatial domain:
\begin{equation}
    \mathcal{K}_{\ell}(\mathbf{z})
    =
    \operatorname{Re}\left[
    \mathcal{F}^{-1}
    \left(
        R_{\ell} \cdot \mathcal{F}(\mathbf{z})
    \right)
    \right],
\end{equation}
where $\mathcal{F}$ and $\mathcal{F}^{-1}$ denote the Fourier transform and inverse transform over the spatial dimensions. The product $R_{\ell} \cdot \mathcal{F}(\mathbf{z})$ denotes mode-wise multiplication with learned channel mixing over the retained Fourier modes. For retained modes, the Fourier weights have shape
\begin{equation}
    R_{\ell} \in \mathbb{C}^{C_z \times C_z \times k_1 \times k_2}.
\end{equation}
In practice, $R_{\ell}$ is applied only to the retained low-frequency modes, while the remaining modes are set to zero before the inverse transform. Since each retained Fourier mode has global spatial support, this branch allows distant spatial locations to interact through low-mode feature components with $O(N\log N)$ scaling for $N=HW$.

The local branch provides spatially localized feature updates:
\begin{equation}
    \mathcal{C}_{\ell}(\mathbf{z}) =
    \mathrm{Conv}_{3\times3}^{\ell}(\mathbf{z}).
\end{equation}
This branch complements the Fourier branch. Low-mode Fourier updates provide global field interaction, but they may suppress fine spatial variations when used alone. The local convolutional branch preserves local feature variations that are important for boundaries, occlusions, and thin structures. The two branches jointly update the same latent feature field before disparity decoding, producing a representation that combines long-range field context with local spatial detail.

\paragraph{Coordinate-conditioned disparity decoder.}
After $L$ hybrid Fourier-local layers, the final latent field $\mathbf{z}^{L}$ is defined on the image grid. For each coordinate $\mathbf{q}\in[-1,1]^2$, we obtain a feature vector by bilinear sampling:
\begin{equation}
    \mathbf{h}(\mathbf{q}) =
    \mathrm{bilinear\_sample}(\mathbf{z}^{L}, \mathbf{q}),
\end{equation}
and concatenate the sampled feature with the coordinate:
\begin{equation}
    \tilde{\mathbf{h}}(\mathbf{q}) =
    [\mathbf{h}(\mathbf{q}); \mathbf{q}].
\end{equation}

A multilayer perceptron maps this coordinate-conditioned feature to the parameters of a Gaussian mixture:
\begin{equation}
    \left(
    \boldsymbol{\pi}(\mathbf{q}),
    \boldsymbol{\mu}(\mathbf{q}),
    \boldsymbol{\sigma}(\mathbf{q})
    \right)
    =
    g_{\theta}\left(\tilde{\mathbf{h}}(\mathbf{q})\right).
\end{equation}
Here, $f_{\theta}$ denotes the full FreqLF model, while $g_{\theta}$ denotes the final MLP decoder. The mixture weights are normalized with a softmax, ensuring
\begin{equation}
    \sum_{k=1}^{K}\pi_k(\mathbf{q}) = 1,
    \qquad
    \pi_k(\mathbf{q}) \geq 0,
\end{equation}
and the standard deviations are parameterized with a softplus function to ensure $\sigma_k(\mathbf{q}) > 0.$

The decoder predicts disparity by bilinearly sampling the learned latent field rather than selecting from a discrete disparity volume. Since the latent representation is computed on an $H\times W$ grid, FreqLF should be interpreted as a coordinate-conditioned decoder over a grid-based latent field, rather than as a fully implicit continuous representation.

\paragraph{Probabilistic output and training objective.}
The predicted mixture parameters define the conditional disparity distribution
\begin{equation}
    p(d \mid \mathbf{q})
    =
    \sum_{k=1}^{K}
    \pi_k(\mathbf{q})
    \mathcal{N}
    \left(
        d \mid \mu_k(\mathbf{q}), \sigma_k^2(\mathbf{q})
    \right),
\end{equation}
where $\pi_k$ are mixture weights, $\mu_k$ are component means, and $\sigma_k$ are component standard deviations. At inference, the final disparity prediction is the mixture mean:
\begin{equation}
    \hat{d}(\mathbf{q})
    =
    \sum_{k=1}^{K}
    \pi_k(\mathbf{q}) \mu_k(\mathbf{q}).
\end{equation}

The model is trained by minimizing the negative log-likelihood over valid ground-truth pixels:
\begin{equation}
    \mathcal{L}_{\mathrm{NLL}}
    =
    -
    \frac{1}{|\mathcal{V}|}
    \sum_{\mathbf{q}_i \in \mathcal{V}}
    \log
    \left[
        \sum_{k=1}^{K}
        \pi_k(\mathbf{q}_i)
        \mathcal{N}
        \left(
            d_i \mid \mu_k(\mathbf{q}_i), \sigma_k^2(\mathbf{q}_i)
        \right)
    \right],
\end{equation}
where $\mathcal{V}\subset\Omega$ denotes the set of normalized coordinates corresponding to valid ground-truth pixels. We use the mixture mean for disparity evaluation, while uncertainty estimates derived from the predicted mixture distribution are used only for diagnostic analysis and are not part of the primary benchmark metric.

\paragraph{Enhanced front-end variant.}
The base FreqLF model uses lightweight EPI streams and avoids explicit cost-volume construction. To test whether the proposed Fourier-local latent field backbone is compatible with stronger angular encoders, we also evaluate an enhanced variant, FreqLF+. In FreqLF+, the lightweight EPI encoder is replaced by an occlusion-aware front-end, while the Fourier-local latent-field backbone and coordinate-conditioned decoder remain unchanged. FreqLF+ is included as a controlled enhanced-front-end variant to test whether the proposed Fourier-local backbone benefits from stronger angular features. It is not used to support the cost-volume-free claim; all comparisons involving cost-volume-free reasoning refer to the base FreqLF model.

\paragraph{Implementation summary.}
Unless otherwise stated, FreqLF uses $C_z=128$ latent channels, $L=4$ hybrid Fourier-local layers, $k_1=k_2=16$ retained Fourier modes, and a $K=5$-component Gaussian mixture decoder. All models are trained end-to-end with Adam and a negative log-likelihood loss. Full architecture, training, and augmentation details are provided in Appendix~\ref{app:implementation}.

\section{Experiments}
\label{sec:experiments}

We evaluate FreqLF on the HCI 4D Light Field Benchmark~\citep{honauer2016dataset}. The experiments address four questions: (i) whether Fourier-local latent field learning approaches strong light-field disparity baselines, (ii) whether the Fourier and local branches provide complementary updates, (iii) whether the base cost-volume-free model scales with spatial resolution, and (iv) where the method succeeds or remains limited beyond average error.

\paragraph{Dataset and protocol.}
We use the four HCI stratified scenes: Backgammon, Dots, Pyramids, and Stripes. We train on the HCI training scenes and evaluate on the held-out stratified scenes. The reported evaluation scenes are not used for training or hyperparameter selection. Each light field contains a $9\times9$ grid of sub-aperture views at $512\times512$ resolution. We estimate central-view disparity and exclude invalid pixels and an $11$-pixel image border. For broader coverage, we also evaluate FreqLF+ on the 16-scene HCI additional split and report the full results in Appendix.

\paragraph{Metrics.}
We report MSE$\times10^2$ and BadPix$(\tau)$:
\begin{equation}
    \mathrm{MSE}\times 10^2
    =
    10^2 \cdot
    \frac{1}{|\mathcal{V}|}
    \sum_{\mathbf{q}_i\in\mathcal{V}}
    \left(\hat d_i-d_i\right)^2 ,
\end{equation}
\begin{equation}
    \mathrm{BadPix}(\tau)
    =
    100\cdot
    \frac{1}{|\mathcal{V}|}
    \sum_{\mathbf{q}_i\in\mathcal{V}}
    \mathbb{1}\left[
        |\hat d_i-d_i|>\tau
    \right],
\end{equation}
where $\mathcal{V}$ is the set of valid pixels. MSE emphasizes large squared errors, while BadPix$(0.03)$ and BadPix$(0.07)$ measure strict and more tolerant error rates.

\paragraph{Model variants.}
FreqLF denotes the base model with an EPI-guided feature encoder, hybrid Fourier-local layers, and a coordinate-conditioned Gaussian-mixture decoder. FreqLF+ keeps the same Fourier-local backbone and decoder but replaces the lightweight EPI encoder with an enhanced EPI front-end. FreqLF+ tests whether the proposed latent field backbone benefits from stronger angular encoding; cost-volume-free claims refer only to the base FreqLF model.

\begin{table}[t]
  \caption{MSE$\times10^2$ on the four HCI stratified scenes. Lower is better. Bold indicates the best or near-best values per column. Rank and Gap are computed from the average MSE, with Gap measuring the relative increase over the best average score.}
  \label{tab:mse_stratified_large}
  \centering
  \scriptsize
  \setlength{\tabcolsep}{1.8pt}
  \renewcommand{\arraystretch}{0.90}
  \begin{tabular*}{\textwidth}{@{\extracolsep{\fill}}llp{2.65cm}ccccccc}
    \toprule
    Method & Setting & Representation
    & Backgammon & Dots & Pyramids & Stripes & Average & Rank & Gap \\
    \midrule
    LFAttNet~\citep{tsai2020attention} & Supervised & View attention
    & 3.648 & 1.425 & \textbf{0.004} & 0.892 & \textbf{1.492} & 1 & 0.0\% \\
    SubFocal-L~\citep{chao2023learning} & Supervised & Cost/focal stack
    & 3.868 & \textbf{1.279} & \textbf{0.005} & 0.874 & \textbf{1.507} & 2 & 1.0\% \\
    Epinet-fcn-m~\citep{shin2018epinet} & Supervised & EPI stacks\epimark
    & \textbf{3.705} & 1.475 & 0.007 & 0.932 & \textbf{1.530} & 4 & 2.5\% \\
    DistgDisp~\citep{wang2022disentangling} & Supervised & Disentangled LF features
    & 4.712 & \textbf{1.367} & \textbf{0.004} & 0.917 & 1.750 & 6 & 17.3\% \\
    OBER-cross-ANP~\citep{hci_obercrossanp} & Optimization & Optimization refinement
    & 4.700 & 1.757 & 0.008 & 1.435 & 1.975 & 7 & 32.4\% \\
    FastLFNet~\citep{huang2021fast} & Supervised & Lightweight LF features
    & 3.986 & 3.407 & 0.018 & 0.984 & 2.099 & 8 & 40.7\% \\
    UnLFDISP~\citep{zhou2023beyond} & Unsupervised & Photometric consistency
    & 6.857 & 1.435 & 0.016 & 1.798 & 2.527 & 9 & 69.4\% \\
    OPAL~\citep{li2023opal} & Unsupervised & Occlusion-aware loss
    & 5.177 & 7.442 & 0.016 & 1.362 & 3.499 & 10 & 134.5\% \\
    CAE~\citep{park2017robust} & Optimization & Optimization estimation
    & 6.074 & 5.082 & 0.048 & 3.556 & 3.690 & 11 & 147.3\% \\
    PnPusp~\citep{iwatsuki2022unsupervised} & Unsupervised & Plug-and-play prior
    & 9.399 & 4.769 & 0.021 & 2.371 & 4.140 & 12 & 177.5\% \\
    PS\_RF~\citep{jeon2018depth} & Optimization & Refocused image cues
    & 6.892 & 8.338 & 0.043 & 1.382 & 4.164 & 13 & 179.1\% \\
    SPO~\citep{zhang2016robust} & Traditional & EPI geometry\epimark
    & 4.587 & 5.238 & 0.043 & 6.955 & 4.206 & 14 & 181.9\% \\
    NDF~\citep{shi2025iterative} & Iterative & Continuous neural field
    & 8.929 & 5.433 & 0.016 & 2.518 & 4.224 & 15 & 183.1\% \\
    AMN~\citep{liu2023adaptive} & Optimization & Adaptive matching norm
    & 12.287 & 2.936 & 0.023 & 2.128 & 4.344 & 16 & 191.2\% \\
    EPI ORM~\citep{li2020epi} & Supervised & EPI orientation reg.\epimark
    & \textbf{3.411} & 14.480 & 0.016 & 1.744 & 4.913 & 17 & 229.3\% \\
    OAVC~\citep{han2021novel} & Optimization & Angular consistency
    & \textbf{3.835} & 16.580 & 0.040 & 1.316 & 5.443 & 18 & 264.8\% \\
    LF\_OCC~\citep{wang2015occlusion} & Optimization & Occlusion-aware LF cues
    & 21.587 & 3.301 & 0.098 & 8.131 & 8.279 & 19 & 454.9\% \\
    \midrule
    FreqLF & Supervised & EPI + Fourier-local field\epimark
    & \textbf{3.880} & \textbf{1.460} & \textbf{0.005} & \textbf{0.865} & \textbf{1.553} & 5 & 4.1\% \\
    FreqLF+ & Supervised & Enhanced EPI + Fourier-local\epimark
    & \textbf{3.824} & \textbf{1.445} & \textbf{0.005} & \textbf{0.812} & \textbf{1.522} & 3 & 2.0\% \\
    \bottomrule
    \multicolumn{10}{l}{\footnotesize \textcolor{red}{\ensuremath{^{\dagger}}} EPI-centered representation or method.}
  \end{tabular*}
\end{table}

\paragraph{Comparison with existing methods.}
We compare against traditional, optimization-based, supervised, unsupervised, and iterative light-field disparity methods using reported per-scene HCI values~\citep{chao2023learning,shi2025iterative}. Table~\ref{tab:mse_stratified_large} shows that FreqLF+ ranks third by average MSE, within $2.0\%$ of the best reported average score, while the base FreqLF ranks fifth, within $4.1\%$. This supports our claim that FreqLF approaches strong supervised baselines while replacing explicit disparity-volume construction in the base model with Fourier-local updates over EPI-derived latent fields. Among EPI-centered methods (\epimark), FreqLF+ achieves the best average MSE, and the base FreqLF remains competitive with Epinet-fcn-m while improving substantially over traditional EPI geometry and EPI orientation regularization.

\begin{table}[t]
  \caption{BadPix$(0.03)$ and BadPix$(0.07)$ on the four HCI stratified scenes. Lower is better. Bold indicates the best or near-best values per column.}
  \label{tab:badpix_stratified_large}
  \centering
  \scriptsize
  \setlength{\tabcolsep}{2.0pt}
  \renewcommand{\arraystretch}{0.88}
  \begin{tabular*}{\textwidth}{@{\extracolsep{\fill}}lcccccccccc}
    \toprule
    \multirow{2}{*}{Method}
    & \multicolumn{5}{c}{BadPix$(0.03)$ [\%]}
    & \multicolumn{5}{c}{BadPix$(0.07)$ [\%]} \\
    \cmidrule(lr){2-6}\cmidrule(lr){7-11}
    & Backgammon & Dots & Pyramids & Stripes & Average
    & Backgammon & Dots & Pyramids & Stripes & Average \\
    \midrule
    LFAttNet~\citep{tsai2020attention}
    & 3.98 & 3.01 & 0.49 & 5.42 & 3.23
    & 3.13 & 1.43 & 0.20 & 2.93 & 1.92 \\
    SubFocal-L~\citep{chao2023learning}
    & \textbf{3.65} & \textbf{1.13} & 0.54 & \textbf{2.22} & \textbf{1.89}
    & \textbf{3.08} & \textbf{0.94} & 0.25 & 2.12 & \textbf{1.60} \\
    Epinet-fcn-m\epimark~\citep{shin2018epinet}
    & 5.56 & 9.12 & 0.87 & 2.71 & 4.57
    & 3.50 & 2.49 & 0.16 & 2.46 & 2.15 \\
    DistgDisp~\citep{wang2022disentangling}
    & 10.54 & 4.46 & 0.54 & 6.89 & 5.61
    & 5.82 & 1.83 & \textbf{0.11} & 3.91 & 2.92 \\
    OBER-cross-ANP~\citep{hci_obercrossanp}
    & 4.96 & 37.66 & 1.13 & 9.35 & 13.28
    & 3.41 & 9.74 & 0.36 & 3.07 & 4.15 \\
    FastLFNet~\citep{huang2021fast}
    & 11.41 & 41.11 & 2.19 & 32.60 & 21.83
    & 5.14 & 21.17 & 0.62 & 9.44 & 9.09 \\
    UnLFDISP~\citep{zhou2023beyond}
    & 12.19 & 3.38 & 3.43 & 8.51 & 6.88
    & 5.20 & 1.53 & 0.27 & 8.51 & 3.88 \\
    OPAL~\citep{li2023opal}
    & 17.27 & 68.84 & 2.62 & 25.51 & 28.56
    & 8.83 & 48.31 & 0.43 & 25.51 & 20.77 \\
    CAE~\citep{park2017robust}
    & 4.31 & 42.50 & 7.16 & 16.90 & 17.72
    & 3.92 & 12.40 & 1.68 & 7.87 & 6.47 \\
    PnPusp~\citep{iwatsuki2022unsupervised}
    & 15.11 & 48.08 & 4.16 & 43.37 & 27.68
    & 14.20 & 29.86 & 0.66 & 43.37 & 22.02 \\
    PS\_RF~\citep{jeon2018depth}
    & 13.94 & 17.54 & 6.24 & 5.79 & 10.88
    & 7.14 & 7.98 & \textbf{0.11} & 2.96 & 4.55 \\
    SPO\epimark~\citep{zhang2016robust}
    & 8.64 & 35.06 & 6.26 & 15.46 & 16.36
    & 3.78 & 16.27 & 0.86 & 14.99 & 8.98 \\
    NDF~\citep{shi2025iterative}
    & 16.38 & 4.44 & 2.85 & 8.05 & 7.93
    & 6.66 & 4.44 & 0.13 & 8.05 & 4.82 \\
    AMN~\citep{liu2023adaptive}
    & 13.35 & 12.46 & 5.64 & 7.87 & 9.83
    & 6.91 & 18.43 & 0.45 & 7.79 & 8.40 \\
    EPI ORM\epimark~\citep{li2020epi}
    & 7.24 & 6.21 & 0.49 & 11.94 & 6.47
    & 3.99 & 4.79 & \textbf{0.11} & 6.87 & 3.94 \\
    OAVC~\citep{han2021novel}
    & 5.12 & 17.38 & 9.03 & 19.88 & 12.85
    & 3.12 & 6.19 & 0.83 & 2.90 & 3.26 \\
    LF\_OCC~\citep{wang2015occlusion}
    & 69.62 & 23.55 & 20.50 & 21.88 & 33.89
    & 19.01 & 58.22 & 3.17 & 21.26 & 25.42 \\
    \midrule
    FreqLF\epimark
    & 4.83 & \textbf{1.88} & \textbf{0.44} & 3.76 & 2.73
    & 3.48 & \textbf{1.11} & 0.25 & 2.58 & \textbf{1.86} \\
    FreqLF+\epimark
    & 4.61 & \textbf{1.78} & \textbf{0.43} & \textbf{2.48} & \textbf{2.33}
    & \textbf{3.35} & \textbf{1.02} & 0.24 & \textbf{2.05} & \textbf{1.67} \\
    \bottomrule
    \multicolumn{11}{l}{\footnotesize \textcolor{red}{\ensuremath{^{\dagger}}} EPI-centered representation or method.}
  \end{tabular*}
\end{table}

Table~\ref{tab:badpix_stratified_large} reports BadPix$(0.03)$ and BadPix$(0.07)$ on the same scenes. FreqLF+ obtains near-best average scores at both thresholds and the best Stripes result for BadPix$(0.07)$. Among EPI-centered methods (\epimark), FreqLF+ achieves the best average BadPix scores, while the base FreqLF remains competitive with the strongest EPI-stack baseline. These results suggest that Fourier-local latent field processing is a useful extension of EPI-derived feature encoding.

\paragraph{Ablation study.}
Table~\ref{tab:component_ablation} isolates the contribution of each component. Removing the Fourier branch increases mean MSE from $1.553$ to $1.821$, showing that low-mode global feature interaction improves the latent field. Removing the local CNN branch causes a larger degradation to $6.148$, confirming that spatially local updates are essential for preserving fine disparity detail. SE fusion and the probabilistic head also improve accuracy, but their contributions are smaller than the Fourier-local backbone.

\begin{table}[t]
  \caption{Component ablation. Values are mean MSE$\times10^2$ over four scenes and three seeds.}
  \label{tab:component_ablation}
  \centering
  \scriptsize
  \setlength{\tabcolsep}{4pt}
  \renewcommand{\arraystretch}{0.95}
  \begin{tabular}{lccccc}
    \toprule
    Metric 
    & Full 
    & No Fourier 
    & No CNN 
    & No SE 
    & No prob. head \\
    \midrule
    MSE$\times10^2$ 
    & $\mathbf{1.553 \pm 0.038}$ 
    & $1.821 \pm 0.061$ 
    & $6.148 \pm 0.287$ 
    & $1.694 \pm 0.054$ 
    & $1.762 \pm 0.049$ \\
    $\Delta$MSE 
    & -- 
    & +0.268 
    & +4.595 
    & +0.141 
    & +0.209 \\
    \bottomrule
  \end{tabular}
\end{table}

\begin{table}[t!]
  \caption{Fourier bandwidth and computational scaling. Lower MSE, runtime, and memory are better.}
  \label{tab:spectral_and_scaling}
  \centering
  \scriptsize
  \setlength{\tabcolsep}{4pt}
  \renewcommand{\arraystretch}{0.95}

  \begin{tabular}{lcccc}
    \toprule
    \multicolumn{5}{c}{\textbf{Effect of retained Fourier modes}} \\
    \midrule
    $k_{\max}$ & Backgammon & Dots & Pyramids & Stripes \\
    \midrule
    8 & $4.312 \pm 0.11$ & $1.793 \pm 0.08$ & $0.008 \pm 0.001$ & $1.047 \pm 0.05$ \\
    16 & $\mathbf{3.880 \pm 0.09}$ & $\mathbf{1.460 \pm 0.06}$ & $\mathbf{0.005 \pm 0.001}$ & $\mathbf{0.865 \pm 0.04}$ \\
    32 & $3.964 \pm 0.10$ & $1.512 \pm 0.07$ & $0.006 \pm 0.001$ & $0.903 \pm 0.05$ \\
    \bottomrule
  \end{tabular}

  \vspace{0.7em}

  \begin{tabular}{lcccc}
    \toprule
    \multicolumn{5}{c}{\textbf{Inference time and peak GPU memory}} \\
    \midrule
    Resolution & \multicolumn{2}{c}{FreqLF} & \multicolumn{2}{c}{FreqLF+} \\
    \cmidrule(lr){2-3}\cmidrule(lr){4-5}
     & Time [s] & Mem. [MB] & Time [s] & Mem. [MB] \\
    \midrule
    $256\times256$ & 0.312 & 3,248 & 0.718 & 8,192 \\
    $512\times512$ & 1.367 & 6,144 & 3.184 & 18,432 \\
    $768\times768$ & 3.124 & 11,520 & 7.298 & 33,280 \\
    $1024\times1024$ & 5.876 & 19,456 & 13.712 & 51,712 \\
    \bottomrule
  \end{tabular}
\end{table}

\paragraph{Fourier bandwidth and computational scaling.}
Table~\ref{tab:spectral_and_scaling} shows that $k_{\max}=16$ retained Fourier modes gives the best validation MSE across the four stratified scenes. The same table reports inference time and peak GPU memory at increasing spatial resolutions on a single NVIDIA A100 GPU. The base FreqLF remains below $20$ GB at $1024\times1024$, while FreqLF+ is more expensive due to its enhanced front-end. This highlights the main trade-off: FreqLF+ improves accuracy with stronger angular encoding, whereas the base model provides the cost-volume-free and more scalable formulation.

\begin{figure}
  \centering
  \includegraphics[width=8.5cm]{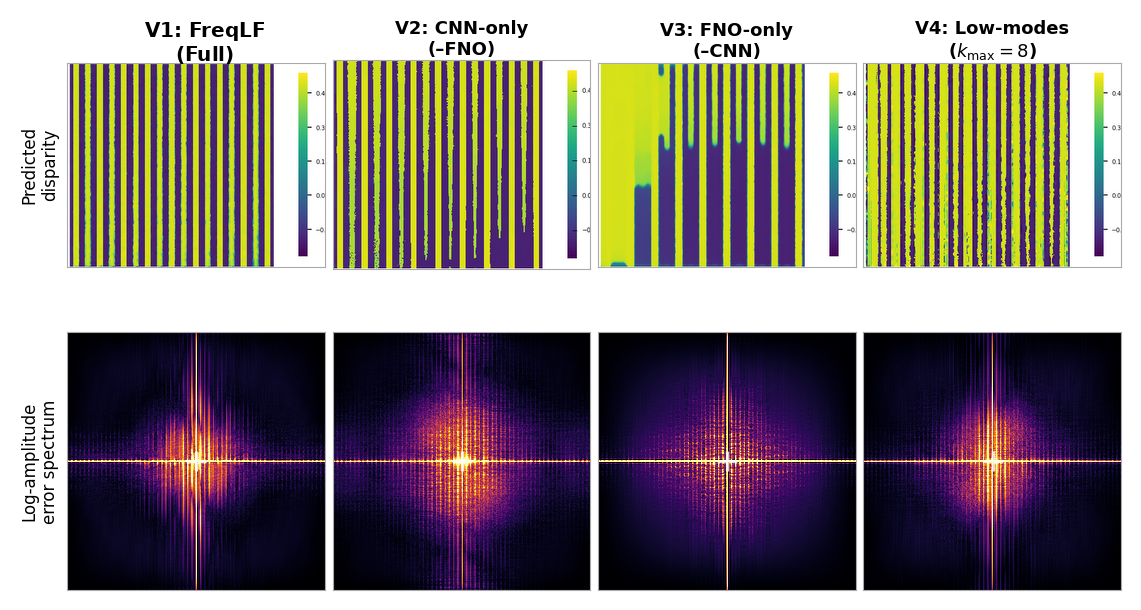}
  \caption{Fourier-domain error analysis on Stripes. The full FreqLF model yields a more balanced spectrum, supporting complementary Fourier and local updates.}
  \label{fig:spectral_error}
\end{figure}

\begin{figure}
  \centering
  \includegraphics[width=8.5cm]{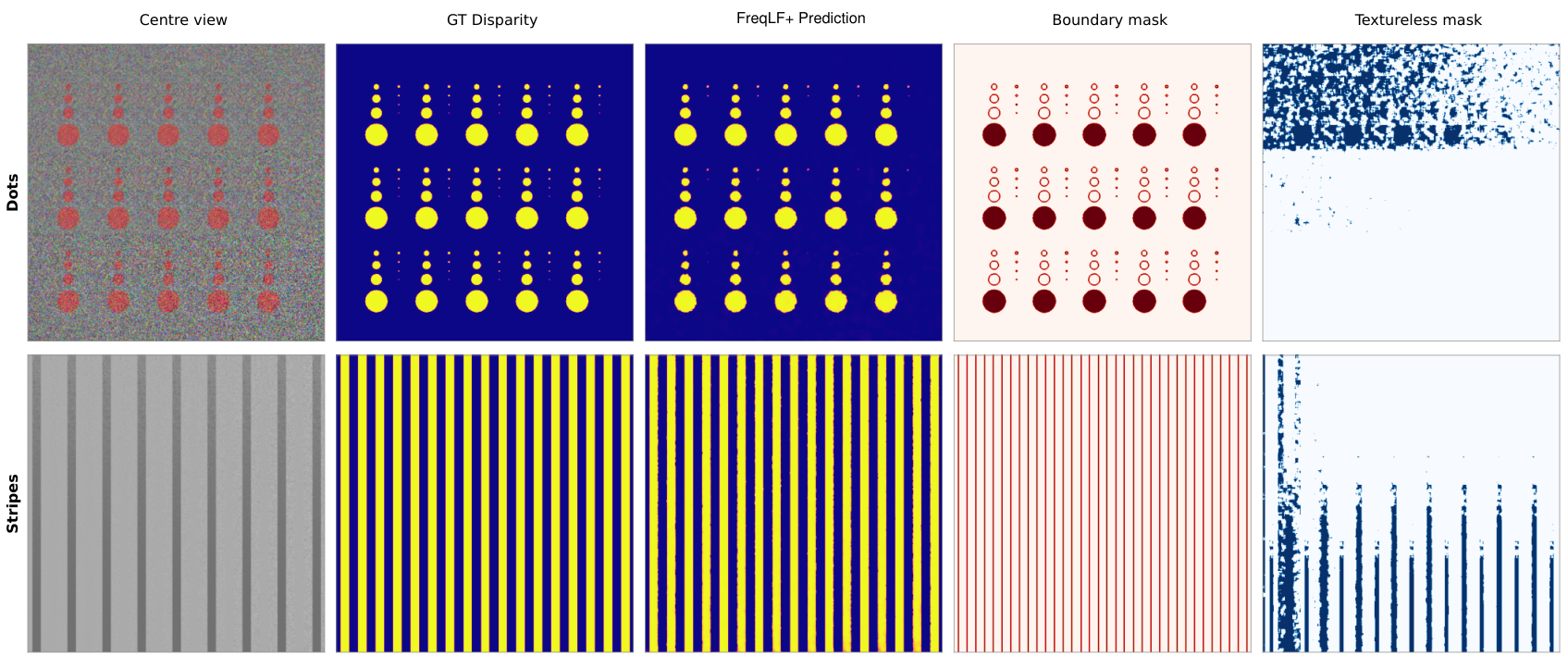}
  \caption{Representative region masks. Boundary and textureless regions are defined by the top $10\%$ ground-truth disparity gradient and bottom $15\%$ local image variance, respectively.}
\label{fig:region_masks_short}
\end{figure}

\paragraph{Frequency and region diagnostics.}
Figure~\ref{fig:spectral_error} analyzes the Fourier-domain error on Stripes. The Fourier-only variant leaves mid-to-high-frequency errors, while the CNN-only variant shows stronger low-frequency bias; the full model gives the most balanced spectrum, supporting complementary Fourier and local updates. Figure~\ref{fig:region_masks_short} shows the masks used for targeted diagnostics. Boundary pixels are the top $10\%$ of ground-truth disparity gradient magnitude, and textureless pixels are the bottom $15\%$ of local central-view variance using a $7\times7$ window. These masks are not official HCI labels, but they help inspect behavior in high-frequency discontinuities and low-texture regions separately.

\section{Conclusion}
\label{sec:conclusion}
We presented FreqLF, an EPI-guided Fourier-local framework for light-field disparity estimation. The main contribution is a field-level alternative to disparity-volume-centered design: instead of explicitly enumerating candidate disparities in the base model, FreqLF predicts disparity from EPI-derived latent features updated by complementary Fourier-domain global interactions and local convolutional operations. Experiments on the HCI 4D Light Field Benchmark show that this formulation approaches strong supervised baselines while retaining a cost-volume-free base architecture. FreqLF+ further improves accuracy with stronger angular encoding, indicating that the proposed Fourier-local backbone can benefit from enhanced EPI features. The MSE and BadPix results, together with branch ablations and scaling analysis, show that the proposed design trades explicit disparity-volume construction for competitive global-local latent field processing. Additional implementation details, extended results, diagnostic analyses, and limitations are provided in the appendix.

\begin{ack}
Do {\bf not} include this section in the anonymized submission, only in the final paper. You can use the \texttt{ack} environment provided in the style file to automatically hide this section in the anonymized submission.
\end{ack}

\bibliographystyle{unsrtnat}
\bibliography{references}

@incollection{levoy2023light,
  title={Light field rendering},
  author={Levoy, Marc and Hanrahan, Pat},
  booktitle={Seminal Graphics Papers: Pushing the Boundaries, Volume 2},
  pages={441--452},
  year={2023},
  publisher={ACM}
}

@incollection{gortler2023lumigraph,
  title={The lumigraph},
  author={Gortler, Steven J and Grzeszczuk, Radek and Szeliski, Richard and Cohen, Michael F},
  booktitle={Seminal Graphics Papers: Pushing the Boundaries, Volume 2},
  pages={453--464},
  year={2023},
  publisher={ACM}
}

@inproceedings{bolles1987epipolar,
  title={Epipolar-Plane Image Analysis: An Approach to Determining Structure from Motion},
  author={Bolles, Robert C. and Baker, H. Harlyn and Marimont, David H.},
  booktitle={International Journal of Computer Vision},
  volume={1},
  pages={7--55},
  year={1987}
}

@inproceedings{honauer2016dataset,
  title={A Dataset and Evaluation Methodology for Depth Estimation on 4D Light Fields},
  author={Honauer, Katrin and Johannsen, Ole and Kondermann, Daniel and Goldluecke, Bastian},
  booktitle={Asian Conference on Computer Vision},
  pages={19--34},
  year={2016},
  organization={Springer}
}

@article{zhang2016robust,
  title={Robust Depth Estimation for Light Field via Spinning Parallelogram Operator},
  author={Zhang, Shuo and Sheng, Hao and Li, Chao and Zhang, Jun and Xiong, Zhang},
  journal={Computer Vision and Image Understanding},
  volume={145},
  pages={148--159},
  year={2016},
  publisher={Elsevier}
}

@inproceedings{shin2018epinet,
  title={EPINET: A Fully-Convolutional Neural Network Using Epipolar Geometry for Depth from Light Field Images},
  author={Shin, Changha and Jeon, Hae-Gon and Yoon, Youngjin and Kweon, In So and Kim, Seon Joo},
  booktitle={Proceedings of the IEEE Conference on Computer Vision and Pattern Recognition},
  pages={4748--4757},
  year={2018}
}

@inproceedings{tsai2020attention,
  title={Attention-Based View Selection Networks for Light-Field Disparity Estimation},
  author={Tsai, Yu-Ju and Liu, Yu-Lun and Ouhyoung, Ming and Chuang, Yung-Yu},
  booktitle={Proceedings of the AAAI Conference on Artificial Intelligence},
  volume={34},
  pages={12095--12103},
  year={2020}
}

@inproceedings{wang2022occlusion,
  title={Occlusion-Aware Cost Constructor for Light Field Depth Estimation},
  author={Wang, Yingqian and Wang, Longguang and Liang, Zhengyu and Yang, Jungang and An, Wei and Guo, Yulan},
  booktitle={Proceedings of the IEEE/CVF Conference on Computer Vision and Pattern Recognition},
  pages={19809--19818},
  year={2022}
}

@article{wang2022disentangling,
  title={Disentangling light fields for super-resolution and disparity estimation},
  author={Wang, Yingqian and Wang, Longguang and Wu, Gaochang and Yang, Jungang and An, Wei and Yu, Jingyi and Guo, Yulan},
  journal={IEEE Transactions on Pattern Analysis and Machine Intelligence},
  volume={45},
  pages={425--443},
  year={2022},
  publisher={IEEE}
}

@article{chao2023learning,
  title={Learning Sub-Pixel Disparity Distribution for Light Field Depth Estimation},
  author={Chao, Wentao and Wang, Xuechun and Wang, Yingqian and Wang, Guanghui and Duan, Fuqing},
  journal={IEEE Transactions on Computational Imaging},
  volume={9},
  pages={1181--1192},
  year={2023},
  publisher={IEEE},
  doi={10.1109/TCI.2023.3336184}
}

@article{li2020fourier,
  title={Fourier neural operator for parametric partial differential equations},
  author={Li, Zongyi and Kovachki, Nikola and Azizzadenesheli, Kamyar and Liu, Burigede and Bhattacharya, Kaushik and Stuart, Andrew and Anandkumar, Anima},
  journal={arXiv preprint arXiv:2010.08895},
  year={2020}
}

@phdthesis{ng2005light,
  title={Light field photography with a hand-held plenoptic camera},
  author={Ng, Ren and Levoy, Marc and Br{\'e}dif, Mathieu and Duval, Gene and Horowitz, Mark and Hanrahan, Pat},
  year={2005},
  school={Stanford university}
}

@article{adelson1992single,
  title={Single lens stereo with a plenoptic camera},
  author={Adelson, Edward H and Wang, John YA},
  journal={IEEE transactions on pattern analysis and machine intelligence},
  volume={14},
  number={2},
  pages={99--106},
  year={1992}
}

@inproceedings{wanner2012globally,
  title={Globally consistent depth labeling of 4D light fields},
  author={Wanner, Sven and Goldluecke, Bastian},
  booktitle={2012 IEEE conference on computer vision and pattern recognition},
  pages={41--48},
  year={2012},
  organization={IEEE}
}

@inproceedings{tao2013depth,
  title={Depth from combining defocus and correspondence using light-field cameras},
  author={Tao, Michael W and Hadap, Sunil and Malik, Jitendra and Ramamoorthi, Ravi},
  booktitle={Proceedings of the IEEE international conference on computer vision},
  pages={673--680},
  year={2013}
}

@inproceedings{hassan2022light,
  title={Light-weight epinet architecture for fast light field disparity estimation},
  author={Hassan, Ali and Sj{\"o}str{\"o}m, M{\aa}rten and Zhang, Tingting and Egiazarian, Karen},
  booktitle={2022 IEEE 24th International Workshop on Multimedia Signal Processing (MMSP)},
  pages={1--5},
  year={2022},
  organization={IEEE}
}

@article{gao2022epi,
  title={EPI light field depth estimation based on a directional relationship model and multiviewpoint attention mechanism},
  author={Gao, Ming and Deng, Huiping and Xiang, Sen and Wu, Jin and He, Zeyang},
  journal={Sensors},
  volume={22},
  number={16},
  pages={6291},
  year={2022},
  publisher={MDPI}
}

@article{liu2022cascade,
  title={Cascade light field disparity estimation network based on unsupervised deep learning},
  author={Liu, Bo and Chen, Jing and Leng, Zhen and Tong, Yanfeng and Wang, Yongtian},
  journal={Optics Express},
  volume={30},
  number={14},
  pages={25130--25146},
  year={2022},
  publisher={Optica Publishing Group}
}

@article{li2023opal,
  title={Opal: Occlusion pattern aware loss for unsupervised light field disparity estimation},
  author={Li, Peng and Zhao, Jiayin and Wu, Jingyao and Deng, Chao and Han, Yuqi and Wang, Haoqian and Yu, Tao},
  journal={IEEE Transactions on Pattern Analysis and Machine Intelligence},
  volume={46},
  number={2},
  pages={681--694},
  year={2023},
  publisher={IEEE}
}

@inproceedings{gao2025epipolar,
  title={Epipolar Consistent Attention Aggregation Network for Unsupervised Light Field Disparity Estimation},
  author={Gao, Chen and Zhang, Shuo and Lin, Youfang},
  booktitle={Proceedings of the IEEE/CVF International Conference on Computer Vision},
  pages={6488--6497},
  year={2025}
}

@article{shi2025iterative,
  title={Iterative approach to reconstructing neural disparity fields from light-field data},
  author={Shi, Ligen and Liu, Chang and Zhao, Xing and Qiu, Jun},
  journal={IEEE Transactions on Computational Imaging},
  volume={11},
  pages={410--420},
  year={2025},
  publisher={IEEE}
}

@article{mildenhall2021nerf,
  title={Nerf: Representing scenes as neural radiance fields for view synthesis},
  author={Mildenhall, Ben and Srinivasan, Pratul P and Tancik, Matthew and Barron, Jonathan T and Ramamoorthi, Ravi and Ng, Ren},
  journal={Communications of the ACM},
  volume={65},
  number={1},
  pages={99--106},
  year={2021},
  publisher={ACM New York, NY, USA}
}

@article{sitzmann2020implicit,
  title={Implicit neural representations with periodic activation functions},
  author={Sitzmann, Vincent and Martel, Julien and Bergman, Alexander and Lindell, David and Wetzstein, Gordon},
  journal={Advances in neural information processing systems},
  volume={33},
  pages={7462--7473},
  year={2020}
}

@article{kovachki2023neural,
  title={Neural operator: Learning maps between function spaces with applications to pdes},
  author={Kovachki, Nikola and Li, Zongyi and Liu, Burigede and Azizzadenesheli, Kamyar and Bhattacharya, Kaushik and Stuart, Andrew and Anandkumar, Anima},
  journal={Journal of Machine Learning Research},
  volume={24},
  number={89},
  pages={1--97},
  year={2023}
}

@article{wanner2013variational,
  title={Variational light field analysis for disparity estimation and super-resolution},
  author={Wanner, Sven and Goldluecke, Bastian},
  journal={IEEE transactions on pattern analysis and machine intelligence},
  volume={36},
  number={3},
  pages={606--619},
  year={2013},
  publisher={IEEE}
}

@inproceedings{chen2021attention,
  title={Attention-based multi-level fusion network for light field depth estimation},
  author={Chen, Jiaxin and Zhang, Shuo and Lin, Youfang},
  booktitle={Proceedings of the AAAI conference on artificial intelligence},
  volume={35},
  pages={1009--1017},
  year={2021}
}

@inproceedings{wang2023epi,
  title={EPI-guided cost construction network for light field disparity estimation},
  author={Wang, Tun and Chen, Rongshan and Cong, Ruixuan and Yang, Da and Cui, Zhenglong and Li, Fangping and Sheng, Hao},
  booktitle={Proceedings of the IEEE/CVF Conference on Computer Vision and Pattern Recognition},
  pages={3438--3446},
  year={2023}
}

@inproceedings{huang2021fast,
  title={Fast light-field disparity estimation with multi-disparity-scale cost aggregation},
  author={Huang, Zhicong and Hu, Xuemei and Xue, Zhou and Xu, Weizhu and Yue, Tao},
  booktitle={Proceedings of the IEEE/CVF international conference on computer vision},
  pages={6320--6329},
  year={2021}
}

@article{iwatsuki2022unsupervised,
  title={Unsupervised disparity estimation from light field using plug-and-play weighted warping loss},
  author={Iwatsuki, Taisei and Takahashi, Keita and Fujii, Toshiaki},
  journal={Signal Processing: Image Communication},
  volume={107},
  pages={116764},
  year={2022},
  publisher={Elsevier}
}

@article{han2021novel,
  title={A novel occlusion-aware vote cost for light field depth estimation},
  author={Han, Kang and Xiang, Wei and Wang, Eric and Huang, Tao},
  journal={IEEE transactions on pattern analysis and machine intelligence},
  volume={44},
  number={11},
  pages={8022--8035},
  year={2021},
  publisher={IEEE}
}

@article{li2020epi,
  title={Epi-based oriented relation networks for light field depth estimation},
  author={Li, Kunyuan and Zhang, Jun and Sun, Rui and Zhang, Xudong and Gao, Jun},
  journal={arXiv preprint arXiv:2007.04538},
  year={2020}
}

@article{park2017robust,
  title={Robust light field depth estimation using occlusion-noise aware data costs},
  author={Park, In Kyu and Lee, Kyoung Mu and others},
  journal={IEEE transactions on pattern analysis and machine intelligence},
  volume={40},
  number={10},
  pages={2484--2497},
  year={2017},
  publisher={IEEE}
}

@article{jeon2018depth,
  title={Depth from a light field image with learning-based matching costs},
  author={Jeon, Hae-Gon and Park, Jaesik and Choe, Gyeongmin and Park, Jinsun and Bok, Yunsu and Tai, Yu-Wing and Kweon, In So},
  journal={IEEE transactions on pattern analysis and machine intelligence},
  volume={41},
  number={2},
  pages={297--310},
  year={2018},
  publisher={IEEE}
}

@inproceedings{wang2015occlusion,
  title={Occlusion-Aware Depth Estimation Using Light-Field Cameras},
  author={Wang, Ting-Chun and Efros, Alexei A. and Ramamoorthi, Ravi},
  booktitle={Proceedings of the IEEE International Conference on Computer Vision},
  pages={3487--3495},
  year={2015}
}

@misc{hci_obercrossanp,
  title={{OBER-cross+ANP}: Occluded Bilateral EPI Regularization with Adaptive Filter Strength, Normal, and Plane Regularization},
  year={2026},
  author={{HCI 4D Light Field Benchmark}},
  howpublished={\url{https://lightfield-analysis.uni-konstanz.de/benchmark/algorithm/obercrossanp}},
  note={Accessed 2026-05-04}
}

@article{zhou2023beyond,
  title={Beyond photometric consistency: Geometry-based occlusion-aware unsupervised light field disparity estimation},
  author={Zhou, Wenhui and Lin, Lili and Hong, Yongjie and Li, Qiujian and Shen, Xingfa and Kuruoglu, Ercan Engin},
  journal={IEEE Transactions on Neural Networks and Learning Systems},
  volume={35},
  number={11},
  pages={15660--15674},
  year={2023},
  publisher={IEEE}
}

@article{liu2023adaptive,
  title={Adaptive matching norm based disparity estimation from light field data},
  author={Liu, Chang and Shi, Ligen and Zhao, Xing and Qiu, Jun},
  journal={Signal Processing},
  volume={209},
  pages={109042},
  year={2023},
  publisher={Elsevier}
}

\appendix
\section{Technical Appendices and Supplementary Material}

\subsection{Detailed Architecture}
\label{app:detailed_architecture}

The main paper presents the conceptual overview of FreqLF. Here, we provide stage-wise architectural details for reproducibility and to clarify the relation between the base model and the enhanced variant. FreqLF consists of three stages: an EPI-guided feature encoder, a hybrid Fourier-local operator backbone, and a coordinate-conditioned probabilistic decoder. FreqLF+ replaces the lightweight EPI encoder with an enhanced occlusion-aware angular front-end, while keeping the hybrid Fourier-local backbone and coordinate-conditioned decoder unchanged.

\begin{figure}[ht]
  \centering
  \includegraphics[width=\linewidth]{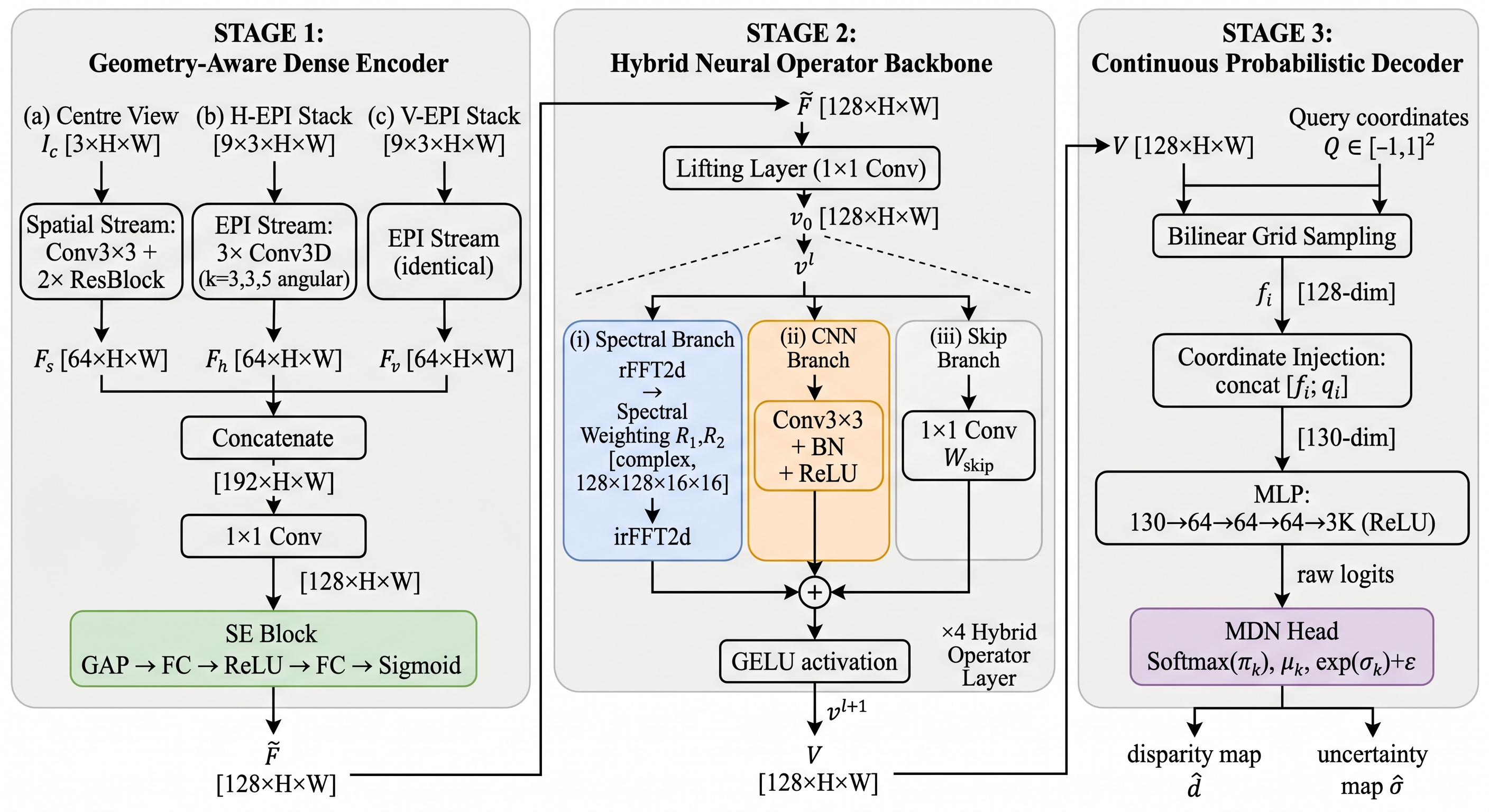}
  \caption{Detailed FreqLF architecture. The model contains an EPI-guided feature encoder, stacked hybrid Fourier-local operator layers, and a coordinate-conditioned decoder.}
  \label{fig:app_full_architecture}
\end{figure}

\begin{figure}[ht]
  \centering
  \includegraphics[width=0.7\linewidth]{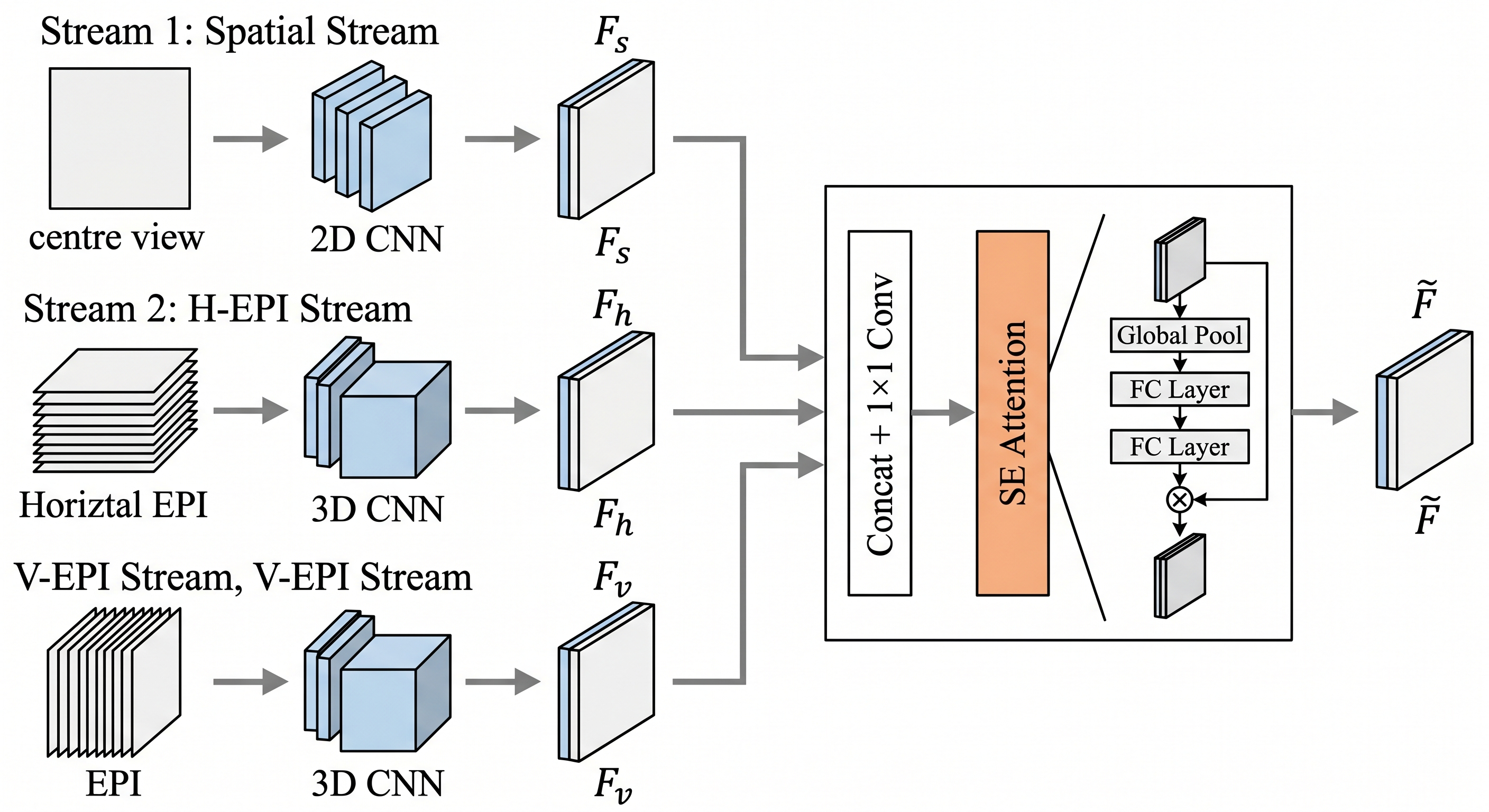}
  \caption{EPI-guided feature encoder. Central-view appearance features are fused with horizontal and vertical EPI features.}
  \label{fig:app_epi_encoder}
\end{figure}

\begin{figure}[ht]
  \centering
  \includegraphics[width=0.6\linewidth]{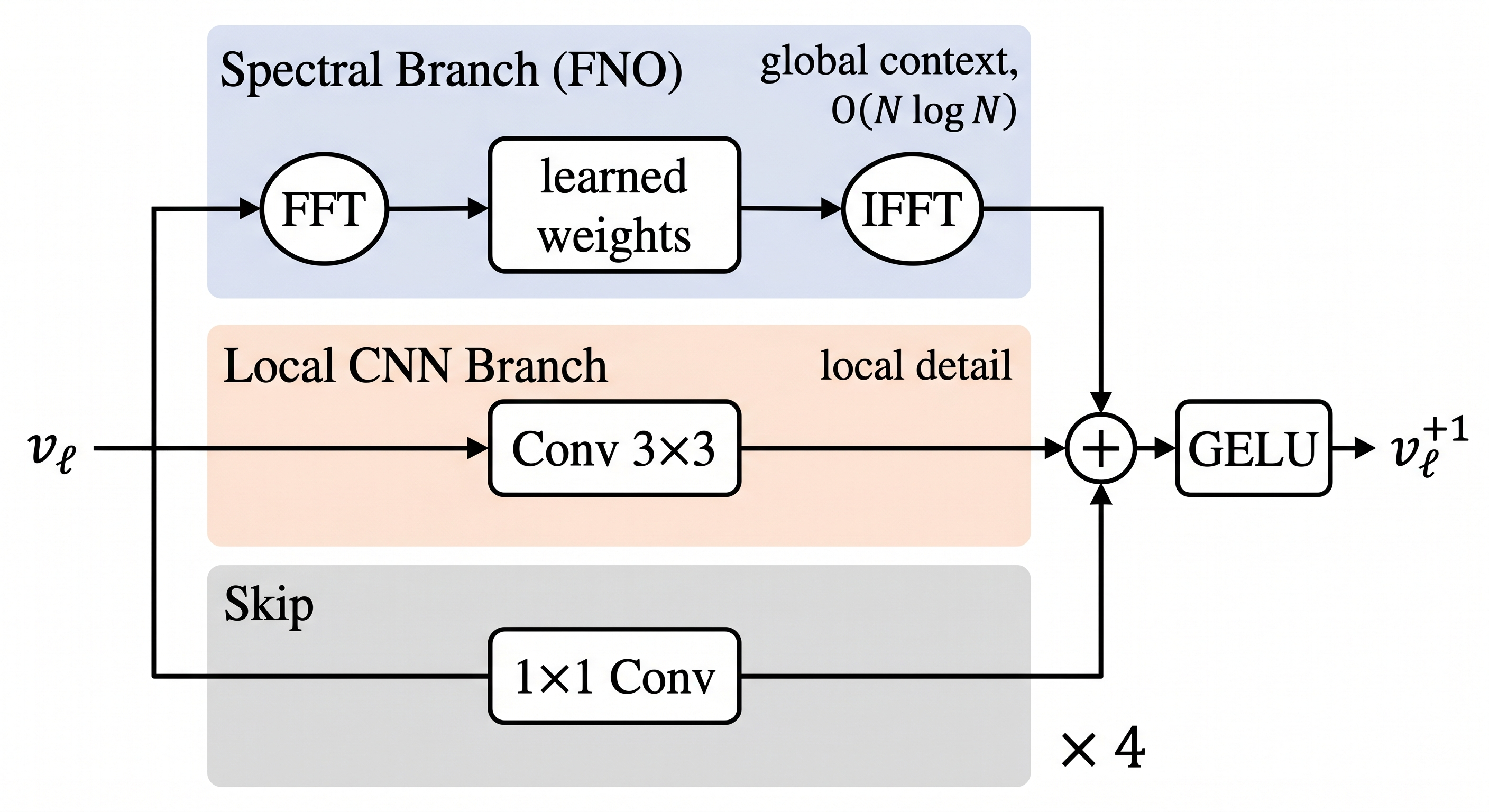}
  \caption{Hybrid Fourier-local operator layer. The Fourier branch provides global low-frequency feature interaction, the local branch preserves spatial detail, and the skip path stabilizes the update.}
  \label{fig:app_hybrid_operator}
\end{figure}

\begin{figure}[ht]
  \centering
  \includegraphics[width=0.65\linewidth]{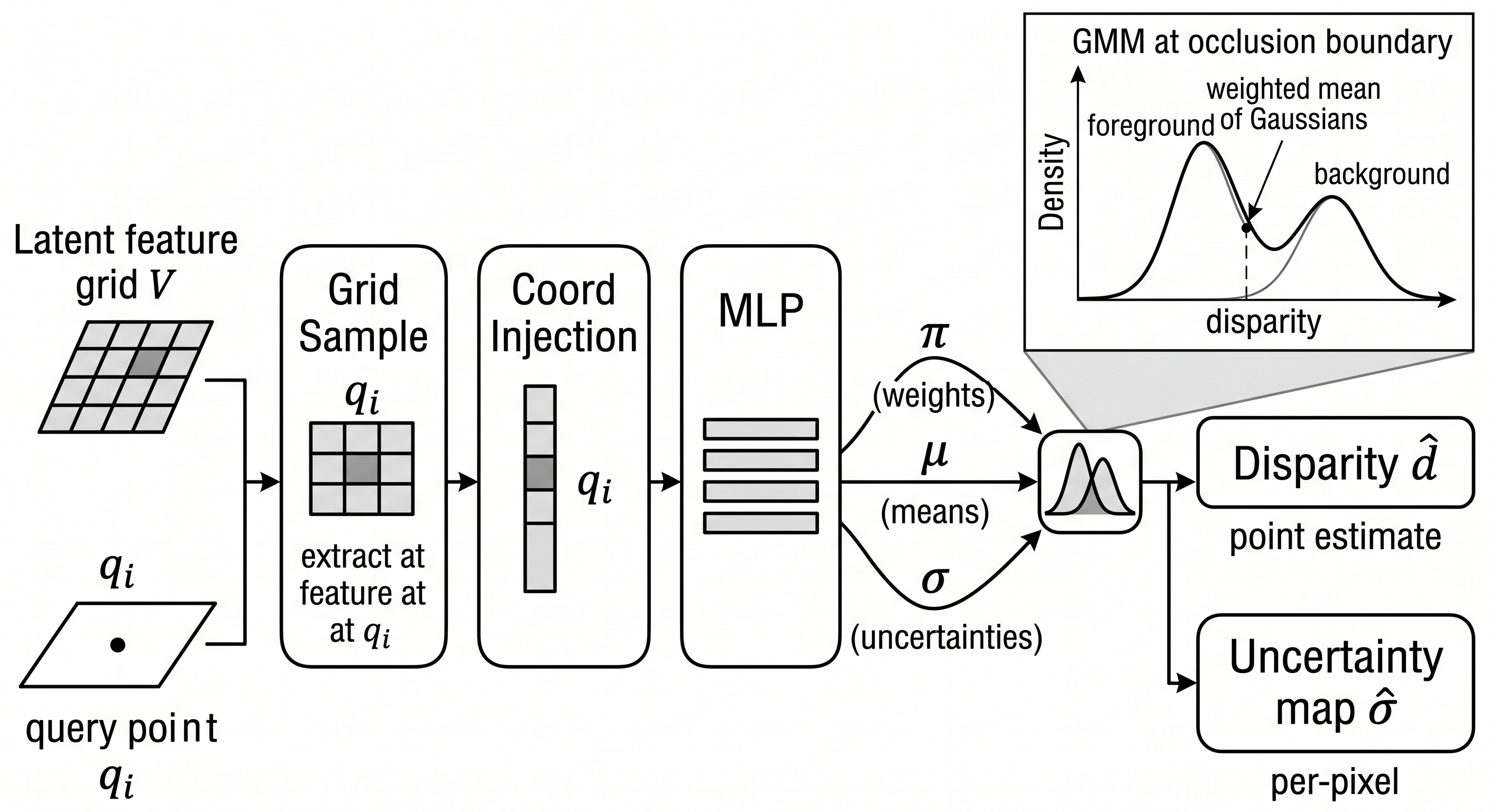}
  \caption{Coordinate-conditioned probabilistic decoder. The latent grid is queried by bilinear sampling and decoded into Gaussian-mixture parameters for disparity prediction.}
  \label{fig:app_decoder}
\end{figure}

\paragraph{EPI-guided feature encoder.}
The base FreqLF encoder processes three input streams: the central view, the horizontal EPI stack, and the vertical EPI stack. The central-view stream extracts spatial appearance features using 2D convolutions. The horizontal and vertical EPI streams use 3D convolutions to aggregate angular information while preserving the spatial grid. These angular streams encode EPI line-slope patterns, which provide geometric cues for disparity. The resulting feature maps are concatenated, projected to the latent channel dimension, and reweighted using channel-wise attention. This produces an EPI-conditioned spatial feature field used by the hybrid operator backbone.

\paragraph{Hybrid Fourier-local backbone.}
The fused feature map is lifted to a latent field with \(C=128\) channels. The backbone uses \(L=4\) hybrid Fourier-local layers. Each layer contains a Fourier branch, a local \(3\times3\) convolutional branch, and a \(1\times1\) residual projection. The Fourier branch applies a 2D FFT over the spatial dimensions, retains \(k_1=k_2=16\) low-frequency modes, applies learned complex-valued weights with channel mixing, and maps the result back to the spatial domain with an inverse FFT. The local branch applies a spatial \(3\times3\) convolution to the same latent field. All layers operate at the full spatial resolution without downsampling, which preserves spatial detail while allowing global field-level interaction.

\paragraph{Coordinate-conditioned probabilistic decoder.}
The final latent grid is queried at normalized image coordinates by bilinear sampling. The sampled latent feature is concatenated with the query coordinate and passed to an MLP decoder. In the probabilistic version, the decoder outputs the parameters of a \(K=5\)-component Gaussian mixture model. The final disparity prediction is the mixture mean. Since the latent representation remains an \(H\times W\) grid, FreqLF should be interpreted as a coordinate-conditioned decoder over a grid-based latent field, rather than as a fully implicit continuous scene representation.

\paragraph{FreqLF+.}
FreqLF+ is an enhanced front-end variant. It replaces the lightweight EPI encoder with an occlusion-aware front-end, while keeping the hybrid Fourier-local backbone, Fourier mode configuration, decoder, and training objective unchanged. We use this variant only as a stronger angular feature extractor to test whether the proposed Fourier-local backbone benefits from improved angular evidence. Therefore, the cost-volume-free claim refers to the base FreqLF model, while FreqLF+ is reported as a modular enhanced variant.

\begin{algorithm}[ht]
\caption{FreqLF forward pass}
\label{alg:freqlf}
\Input{Central view \(I_c \in \mathbb{R}^{3 \times H \times W}\),
       horizontal EPI stack \(\mathbf{E}_h \in \mathbb{R}^{9 \times 3 \times H \times W}\),
       vertical EPI stack \(\mathbf{E}_v \in \mathbb{R}^{9 \times 3 \times H \times W}\),
       query coordinates \(\mathbf{Q} \in \mathbb{R}^{N_q \times 2}\)}
\Output{Mixture weights \(\boldsymbol{\pi} \in \mathbb{R}^{N_q \times K}\),
        means \(\boldsymbol{\mu} \in \mathbb{R}^{N_q \times K}\),
        standard deviations \(\boldsymbol{\sigma} \in \mathbb{R}^{N_q \times K}\)}
\BlankLine

\tcp{Stage 1: EPI-guided feature encoding}
\(\mathbf{F}_s \leftarrow \mathrm{SpatialStream}(I_c)\) \tcp*{\(\mathbb{R}^{64 \times H \times W}\)}
\(\mathbf{F}_h \leftarrow \mathrm{EPIStream}(\mathbf{E}_h)\) \tcp*{\(\mathbb{R}^{64 \times H \times W}\)}
\(\mathbf{F}_v \leftarrow \mathrm{EPIStream}(\mathbf{E}_v)\) \tcp*{\(\mathbb{R}^{64 \times H \times W}\)}
\(\mathbf{F}_{cat} \leftarrow [\mathbf{F}_s;\mathbf{F}_h;\mathbf{F}_v]\)\;
\(\mathbf{F} \leftarrow \mathrm{SEFusion}(\mathbf{F}_{cat})\) \tcp*{\(\mathbb{R}^{128 \times H \times W}\)}
\BlankLine

\tcp{Stage 2: hybrid Fourier-local operator}
\(\mathbf{z}^{0} \leftarrow W_{\mathrm{lift}}(\mathbf{F})\) \tcp*{\(1\times1\) projection to latent field}
\For{\(\ell = 0\) \KwTo \(L-1\)}{
  \(\mathbf{s}^{\ell} \leftarrow \mathcal{K}_{\ell}(\mathbf{z}^{\ell})\) \tcp*{Fourier branch}
  \(\mathbf{c}^{\ell} \leftarrow \mathcal{C}_{\ell}(\mathbf{z}^{\ell})\) \tcp*{local \(3\times3\) branch}
  \(\mathbf{r}^{\ell} \leftarrow W_{\ell}(\mathbf{z}^{\ell})\) \tcp*{\(1\times1\) residual projection}
  \(\mathbf{z}^{\ell+1} \leftarrow \mathrm{GELU}\!\left(\mathbf{s}^{\ell}+\mathbf{c}^{\ell}+\mathbf{r}^{\ell}\right)\)\;
}
\BlankLine

\tcp{Stage 3: coordinate-conditioned probabilistic decoder}
\For{each query coordinate \(\mathbf{q}_i \in \mathbf{Q}\)}{
  \(\mathbf{h}_i \leftarrow \mathrm{BilinearSample}(\mathbf{z}^{L}, \mathbf{q}_i)\)\;
  \(\tilde{\mathbf{h}}_i \leftarrow [\mathbf{h}_i;\mathbf{q}_i]\) \tcp*{coordinate injection}
  \((\hat{\boldsymbol{\pi}}_i,\hat{\boldsymbol{\mu}}_i,\hat{\boldsymbol{\rho}}_i) \leftarrow \mathrm{MLP}(\tilde{\mathbf{h}}_i)\)\;
  \(\pi_{ik} \leftarrow \mathrm{Softmax}_k(\hat{\boldsymbol{\pi}}_i)\)\;
  \(\mu_{ik} \leftarrow \hat{\mu}_{ik}\)\;
  \(\sigma_{ik} \leftarrow \mathrm{softplus}(\hat{\rho}_{ik})+\varepsilon\)\;
}
\Return \(\boldsymbol{\pi},\boldsymbol{\mu},\boldsymbol{\sigma}\)
\end{algorithm}

Algorithm~\ref{alg:freqlf} summarizes the forward pass of the base FreqLF model.

\subsection{Implementation Details}
\label{app:implementation}

\paragraph{Model configuration.}
Unless otherwise stated, all FreqLF experiments use the configuration in Table~\ref{tab:app_model_config}. The default Fourier bandwidth is \(k_1=k_2=16\), which is also the best-performing setting in the spectral bandwidth study.

\begin{table}[ht]
  \caption{Default model configuration.}
  \label{tab:app_model_config}
  \centering
  \small
  \setlength{\tabcolsep}{5pt}
  \renewcommand{\arraystretch}{0.95}
  \begin{tabular}{ll}
    \toprule
    Component & Configuration \\
    \midrule
    Latent channels & \(C=128\) \\
    Hybrid operator layers & \(L=4\) \\
    Fourier modes & \(k_1=k_2=16\) \\
    Local branch & \(3\times3\) convolution \\
    Skip branch & \(1\times1\) convolution \\
    Activation & GELU \\
    Gaussian mixture components & \(K=5\) \\
    Query strategy & full \(H\times W\) grid during evaluation \\
    \bottomrule
  \end{tabular}
\end{table}

\paragraph{Training setup.}
All models are trained with Adam using the default momentum parameters \(\beta_1=0.9\), \(\beta_2=0.999\), and \(\epsilon=10^{-8}\). The learning rate is fixed to \(10^{-4}\), with no scheduler. The batch size is 1, and models are trained for 500 epochs. The training loss is the negative log-likelihood of the ground-truth disparity under the predicted Gaussian mixture. During each iteration, the model processes one light field, queries all valid spatial coordinates, computes the NLL loss, and updates all learnable parameters end-to-end.

\begin{table}[ht]
  \caption{Training and system configuration.}
  \label{tab:app_training_config}
  \centering
  \small
  \setlength{\tabcolsep}{5pt}
  \renewcommand{\arraystretch}{0.95}
  \begin{tabular}{ll}
    \toprule
    Component & Specification \\
    \midrule
    GPU & NVIDIA A100, 80 GB HBM2e \\
    CPU & Intel Xeon, 10 cores \\
    System RAM & 64 GB \\
    OS & Linux \\
    Python & 3.11 \\
    PyTorch & 2.3.1 + CUDA 12.1 \\
    Optimizer & Adam \\
    Learning rate & \(1\times10^{-4}\), constant \\
    Batch size & 1 \\
    Training epochs & 500 \\
    Monitoring & optional Weights \& Biases logging \\
    \bottomrule
  \end{tabular}
\end{table}

\paragraph{Data augmentation.}
We use multi-scale random cropping during training. Let \(S=\min(H,W)\). At each iteration, one of the crop strategies in Table~\ref{tab:app_augmentation} is sampled. The same crop is applied to the central view, EPI stacks, disparity map, and valid mask.

\begin{table}[ht]
  \caption{Multi-scale crop augmentation.}
  \label{tab:app_augmentation}
  \centering
  \small
  \setlength{\tabcolsep}{6pt}
  \renewcommand{\arraystretch}{0.95}
  \begin{tabular}{lll}
    \toprule
    Strategy & Probability & Crop size \\
    \midrule
    No crop & 30\% & full resolution \(H\times W\) \\
    Large crop & 35\% & width and height sampled in \([0.7S,S]\) \\
    Small crop & 35\% & width and height sampled in \([128,0.7S]\) \\
    \bottomrule
  \end{tabular}
\end{table}

For ablation experiments, all variants are trained from scratch under identical hyperparameters, changing only the component under study. To reduce wall-clock time, ablations use a forced small-crop setting with crop sizes sampled in \([64,0.7S]\). This keeps the relative comparison controlled while reducing training cost.

\paragraph{Evaluation protocol.}
We follow the HCI 4D Light Field Benchmark protocol. The reported stratified scenes are used only for evaluation and are not used during training or hyperparameter selection. All evaluations use the valid pixel mask provided by the benchmark and exclude an 11-pixel image border. Unless otherwise stated, inference is performed at the native \(512\times512\) resolution with a single model and a single forward pass. No test-time ensembling, multi-model fusion, or post-processing is applied.

\subsection{Additional HCI Results}
\label{app:additional_results}

In addition to the four HCI stratified scenes used in the main paper, we evaluate FreqLF+ on the 16-scene HCI additional split. Table~\ref{tab:app_hci_additional} reports per-scene MSE\(\times10^2\) and BadPix\((0.07)\). These results provide broader coverage across different textures, geometries, and occlusion conditions.

\begin{table}[ht]
  \caption{FreqLF+ on the HCI additional split. Lower is better.}
  \label{tab:app_hci_additional}
  \centering
  \small
  \setlength{\tabcolsep}{8pt}
  \renewcommand{\arraystretch}{0.95}
  \begin{tabular}{lcc}
    \toprule
    Scene & MSE\(\times10^2\) & BadPix\((0.07)\) [\%] \\
    \midrule
    Antinous & 0.460 & 0.672 \\
    Boardgames & 0.032 & 1.220 \\
    Dishes & 0.157 & 0.944 \\
    Greek & 1.569 & 6.238 \\
    Kitchen & 12.596 & 24.371 \\
    Medieval2 & 0.054 & 0.967 \\
    Museum & 2.370 & 6.785 \\
    Pens & 0.355 & 1.264 \\
    Pillows & 0.024 & 0.296 \\
    Platonic & 0.163 & 0.157 \\
    Rosemary & 2.876 & 6.845 \\
    Table & 0.583 & 6.017 \\
    Tomb & 0.025 & 0.292 \\
    Tower & 0.763 & 1.690 \\
    Town & 0.127 & 1.280 \\
    Vinyl & 7.552 & 16.246 \\
    \midrule
    Average & 1.857 & 4.705 \\
    \bottomrule
  \end{tabular}
\end{table}

The additional split shows that FreqLF+ performs strongly on several scenes with clean geometry and moderate texture, such as Boardgames, Pillows, Tomb, and Platonic. The largest errors occur on Kitchen and Vinyl, which contain challenging specular, semi-transparent, or out-of-distribution appearance effects. Moderate errors occur on Rosemary, Museum, and Greek, which contain fine-scale or high-frequency geometry. These results indicate that FreqLF+ generalizes well across many HCI scenes, while complex materials and difficult occlusion patterns remain challenging.

\subsection{Ablation and Diagnostic Protocols}
\label{app:ablation_diagnostics}

This section provides additional details on the ablation and diagnostic protocols used to interpret the Fourier-local design. These analyses are intended to clarify why both branches are included and how the diagnostic masks are constructed.

\paragraph{Component ablations.}
The component ablation in the main paper isolates the effect of the Fourier branch, local CNN branch, SE fusion, and probabilistic head. All ablated variants are trained from scratch using the same optimizer, learning rate, number of epochs, data processing, and evaluation protocol. Only the component under study is changed. This keeps the comparison focused on the contribution of each module.

\paragraph{Interpretation of the Fourier branch.}
The Fourier branch is designed to update the latent field using low-frequency modes with global spatial support. This is useful for disparity regions that require long-range consistency, such as smooth surfaces, planar structures, and textureless areas where local evidence can be weak. Removing the Fourier branch increases the average error, indicating that global low-mode interaction improves the latent field beyond purely local processing.

\paragraph{Interpretation of the local branch.}
The local \(3\times3\) branch is designed to preserve and refine spatial variations. This is especially important near occlusion boundaries, thin structures, and abrupt foreground-background transitions. Removing the local branch leads to the largest degradation in the ablation study, which confirms that low-mode Fourier updates alone are insufficient for accurate disparity estimation. The full model therefore combines global Fourier-domain interaction with spatially localized updates.

\paragraph{Fourier bandwidth.}
The spectral bandwidth study varies the retained Fourier modes \(k_{\max}\). Small bandwidths provide strong smoothing but can under-represent spatial detail. Larger bandwidths increase the number of retained frequency components, but they do not necessarily improve validation accuracy. In our experiments, \(k_{\max}=16\) gives the best balance between global interaction, local detail preservation, and computational cost.

\paragraph{Boundary mask definition.}
Boundary diagnostics are computed from the ground-truth disparity map. Let \(d\) denote the ground-truth disparity and let
\[
G(\mathbf{q}) =
\sqrt{
\left(\partial_x d(\mathbf{q})\right)^2 +
\left(\partial_y d(\mathbf{q})\right)^2
}
\]
denote the disparity-gradient magnitude. Boundary pixels are selected as the top \(10\%\) of valid pixels according to \(G(\mathbf{q})\). These pixels approximate high-frequency disparity transitions such as object boundaries and foreground-background discontinuities.

\paragraph{Textureless mask definition.}
The HCI benchmark does not provide official textureless-region annotations. We therefore define textureless regions diagnostically from the central view. We convert the central view to intensity and compute the local intensity variance in a \(7\times7\) window. Textureless pixels are selected as the bottom \(15\%\) of valid pixels according to this local variance. These pixels approximate regions where local appearance cues are weak and angular or global field information becomes more important.

\paragraph{Use of diagnostic masks.}
The boundary and textureless masks are not official HCI semantic labels. They are used only as diagnostic tools to separate two common light-field disparity failure modes: high-frequency discontinuities and low-texture regions. The masks help evaluate whether the Fourier-local design behaves as intended. The Fourier branch is expected to support globally consistent estimates in weakly textured or smooth regions, while the local branch is expected to improve boundary and thin-structure behavior.

\begin{figure}[ht]
  \centering
  \includegraphics[width=\linewidth]{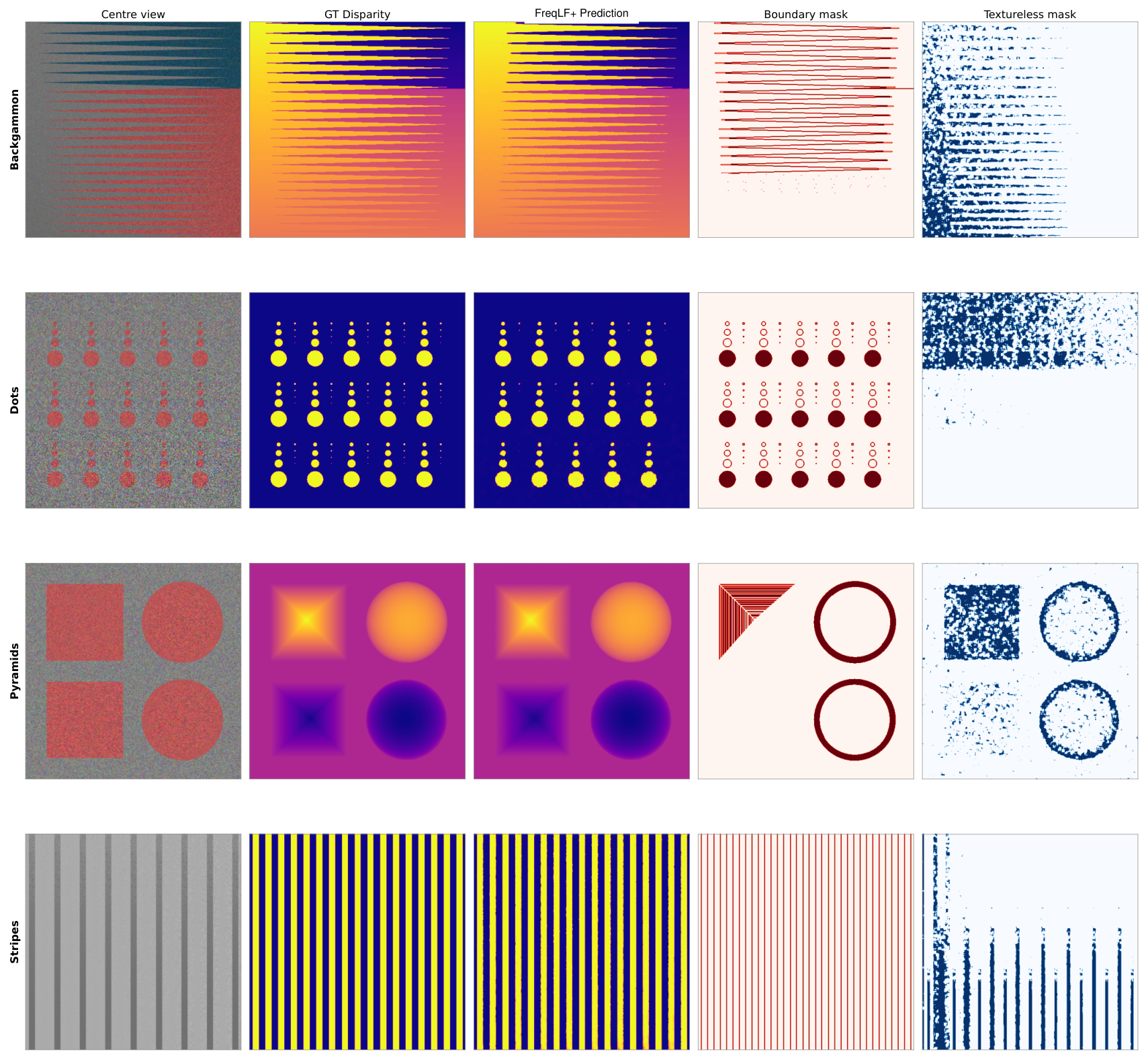}
  \caption{Representative diagnostic masks. Boundary regions are selected from the top \(10\%\) of ground-truth disparity-gradient magnitude, and textureless regions from the bottom \(15\%\) of local central-view intensity variance.}
  \label{fig:app_region_masks}
\end{figure}

\paragraph{Region-specific quantitative results.}
Table~\ref{tab:app_region_specific} reports region-specific MSE\(\times10^2\) and BadPix\((0.07)\) for FreqLF+ on the four HCI stratified scenes. Errors are substantially higher in boundary regions than in textureless regions. The mean boundary error is \(5.254\) MSE\(\times10^2\) and \(11.467\%\) BadPix\((0.07)\), while the mean textureless-region error is \(0.171\) MSE\(\times10^2\) and \(0.601\%\) BadPix\((0.07)\). This suggests that the model handles weak-texture regions well, while sharp disparity transitions remain the main source of residual error.

\begin{table}[ht]
  \caption{Region-specific MSE\(\times10^2\) and BadPix\((0.07)\) for FreqLF+. Boundary and textureless masks are diagnostic masks, not official HCI labels. Lower is better.}
  \label{tab:app_region_specific}
  \centering
  \small
  \setlength{\tabcolsep}{6pt}
  \renewcommand{\arraystretch}{0.95}
  \begin{tabular}{lcccc}
    \toprule
    & \multicolumn{2}{c}{Boundary} & \multicolumn{2}{c}{Textureless} \\
    \cmidrule(lr){2-3}
    \cmidrule(lr){4-5}
    Scene & MSE\(\times10^2\) & BadPix\((0.07)\) [\%] & MSE\(\times10^2\) & BadPix\((0.07)\) [\%] \\
    \midrule
    Backgammon & 9.028 & 12.980 & 0.017 & 0.107 \\
    Dots       & 9.674 & 14.297 & 0.589 & 0.948 \\
    Pyramids   & 0.034 & 1.791  & 0.006 & 0.169 \\
    Stripes    & 2.281 & 16.798 & 0.070 & 1.181 \\
    \midrule
    Mean       & 5.254 & 11.467 & 0.171 & 0.601 \\
    \bottomrule
  \end{tabular}
\end{table}

Backgammon and Dots have the highest boundary errors, reflecting the difficulty of sharp, high-contrast depth edges and sparse repeated structures. Pyramids has the lowest boundary error because its disparity changes are smoother and less dominated by abrupt occlusion boundaries. Textureless regions show consistently low error across all scenes, supporting the view that EPI-derived global field processing is effective when local appearance evidence is weak.

\paragraph{Frequency-domain error analysis.}
For frequency-domain diagnostics, we compute the absolute disparity error map for each model variant and visualize its Fourier magnitude spectrum. This analysis is used to inspect whether errors are concentrated in low-frequency components, high-frequency components, or both. The Fourier-only variant tends to leave mid-to-high-frequency errors, while the CNN-only variant shows stronger low-frequency bias. The full model gives a more balanced error spectrum, supporting the complementary role of the Fourier and local branches.

\subsection{Mathematical Interpretation of the Fourier-Local Update}
\label{app:math_interpretation}

FreqLF does not rely on a formal convergence guarantee or a theoretical optimality proof. However, the hybrid update can be interpreted as a decomposition of the latent-field update into a globally supported low-frequency component and a spatially local component. For a latent field \(\mathbf{z}^{\ell} \in \mathbb{R}^{C \times H \times W}\), each hybrid layer applies
\begin{equation}
\mathbf{z}^{\ell+1}
=
\sigma\!\left(
\mathcal{K}_{\ell}(\mathbf{z}^{\ell})
+
\mathcal{C}_{\ell}(\mathbf{z}^{\ell})
+
W^{1\times1}_{\ell}(\mathbf{z}^{\ell})
\right),
\end{equation}
where \(\mathcal{K}_{\ell}\) is the Fourier-domain branch, \(\mathcal{C}_{\ell}\) is the local convolutional branch, and \(W^{1\times1}_{\ell}\) is a pointwise residual projection.

Ignoring the nonlinearity and channel mixing for interpretation, the Fourier branch applies a learned update on a retained set of low-frequency spatial modes:
\begin{equation}
\mathcal{K}_{\ell}(\mathbf{z})
=
\mathcal{F}^{-1}
\left(
\mathbf{M}_{k}
\odot
\mathbf{R}_{\ell}\mathcal{F}(\mathbf{z})
\right),
\end{equation}
where \(\mathcal{F}\) and \(\mathcal{F}^{-1}\) denote the 2D Fourier transform and inverse transform over the spatial dimensions, \(\mathbf{M}_{k}\) is a binary mask selecting the retained modes, and \(\mathbf{R}_{\ell}\) denotes learned complex-valued mode weights with channel mixing. Since each Fourier basis function has global spatial support, this branch allows distant spatial locations to interact through the retained frequency components. Low-frequency modes are well suited to representing smooth field components such as large planar regions, global disparity trends, and textureless areas.

The local convolutional branch is given by
\begin{equation}
\mathcal{C}_{\ell}(\mathbf{z})
=
\mathrm{Conv}^{3\times3}_{\ell}(\mathbf{z}),
\end{equation}
and has compact spatial support. It therefore provides localized corrections that are important for high-frequency disparity structures, including occlusion boundaries, thin structures, and abrupt foreground-background transitions.

Thus, the hybrid layer can be viewed as a learned spectral-local update. The Fourier branch updates globally supported smooth components of the latent field, while the convolutional branch updates local spatial detail. This interpretation does not imply an exact frequency separation or a formal optimality guarantee, because both branches operate on learned latent features and are followed by nonlinearities. Instead, it motivates the architecture, while the ablation results empirically verify that both branches provide complementary contributions.

\subsection{Uncertainty Diagnostics}
\label{app:uncertainty}

The probabilistic decoder predicts a Gaussian mixture distribution over disparity at each query coordinate. These predictions are used to compute the final disparity estimate and to visualize uncertainty. We use uncertainty only as an auxiliary diagnostic signal, not as a calibrated confidence estimate.

For each query coordinate, the decoder predicts a \(K\)-component Gaussian mixture distribution,
\begin{equation}
p(d\mid \mathbf{q}) =
\sum_{k=1}^{K}
\pi_k(\mathbf{q})
\mathcal{N}
\left(
d \mid \mu_k(\mathbf{q}), \sigma_k^2(\mathbf{q})
\right).
\end{equation}
The point disparity prediction is the mixture mean,
\begin{equation}
\hat d(\mathbf{q}) =
\sum_{k=1}^{K}
\pi_k(\mathbf{q})\mu_k(\mathbf{q}).
\end{equation}
For visualization, we compute the predictive variance using the law of total variance:
\begin{equation}
\mathrm{Var}(d\mid\mathbf{q})
=
\sum_{k=1}^{K}
\pi_k(\mathbf{q})\sigma_k^2(\mathbf{q})
+
\sum_{k=1}^{K}
\pi_k(\mathbf{q})\mu_k^2(\mathbf{q})
-
\left(
\sum_{k=1}^{K}
\pi_k(\mathbf{q})\mu_k(\mathbf{q})
\right)^2 .
\end{equation}

\begin{figure}[ht]
  \centering
  \includegraphics[width=12cm]{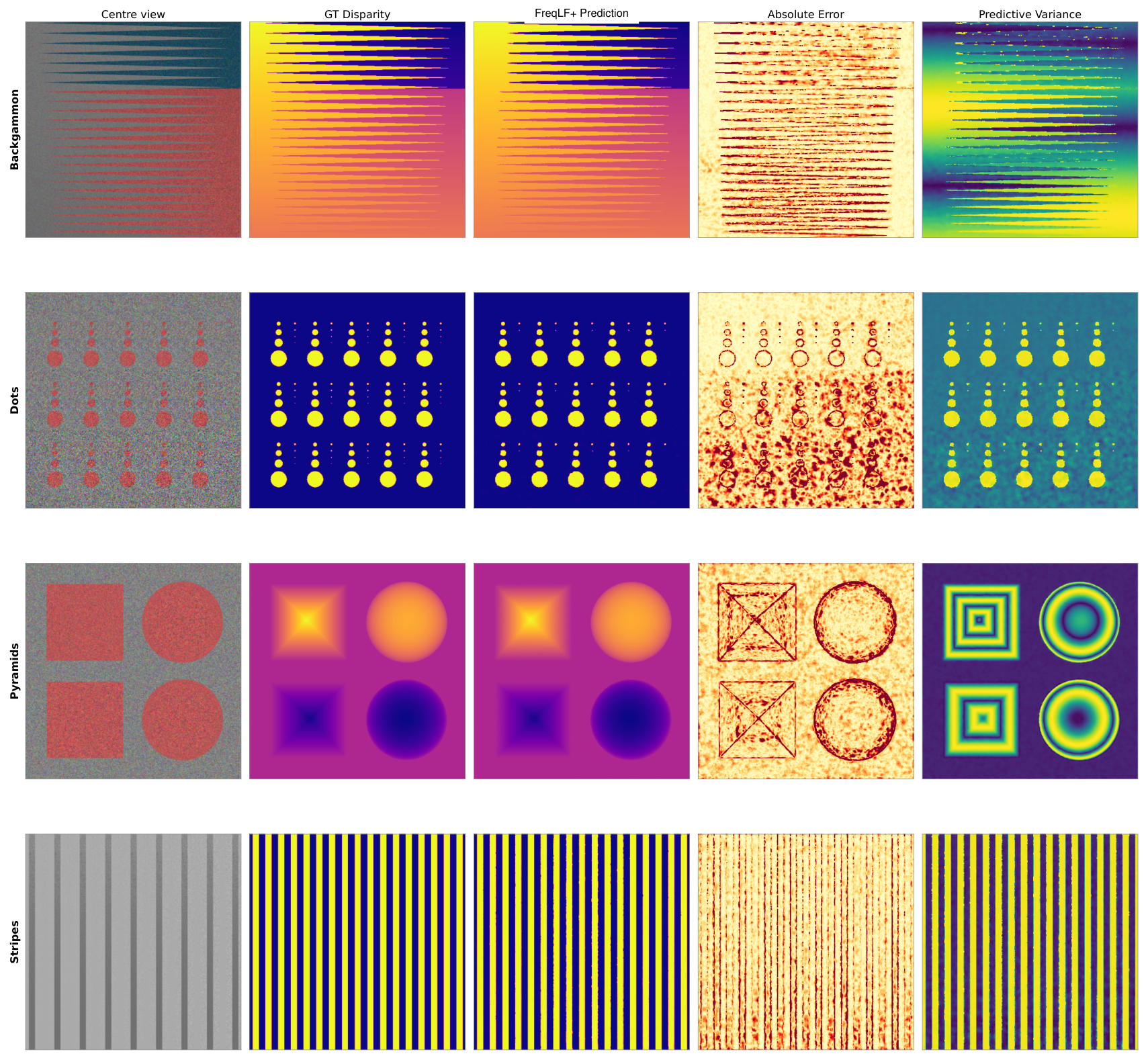}
  \caption{Qualitative uncertainty diagnostics on the HCI stratified scenes.}
  \label{fig:app_uncertainty_grid}
\end{figure}

The uncertainty maps provide useful qualitative diagnostics, especially near boundaries, thin structures, and ambiguous regions. However, qualitative uncertainty maps alone do not imply calibrated confidence. We therefore additionally evaluate the relation between predicted variance and actual disparity error.

\paragraph{Uncertainty-error correlation.}
Table~\ref{tab:app_uncertainty_corr} reports the Spearman and Pearson correlations between predicted variance and absolute disparity error. The pooled Spearman correlation is \(0.069\), and the pooled Pearson correlation is \(0.218\), indicating that predicted variance is only weakly correlated with the actual error. Scene-wise correlations are also inconsistent, including negative Spearman correlations for Backgammon and Stripes.

\begin{table}[ht]
  \caption{Correlation between predicted variance and absolute disparity error for FreqLF+. Higher values would indicate better uncertainty-error alignment.}
  \label{tab:app_uncertainty_corr}
  \centering
  \small
  \setlength{\tabcolsep}{8pt}
  \renewcommand{\arraystretch}{0.95}
  \begin{tabular}{lcc}
    \toprule
    Scene & Spearman \(\rho\) & Pearson \(r\) \\
    \midrule
    Backgammon & -0.037 & 0.281 \\
    Dots       & 0.486  & 0.288 \\
    Pyramids   & 0.254  & 0.042 \\
    Stripes    & -0.156 & 0.113 \\
    \midrule
    Pooled     & 0.069  & 0.218 \\
    \bottomrule
  \end{tabular}
\end{table}

\paragraph{Coverage-based calibration.}
We also evaluate coverage-based calibration. For each confidence level \(p\), we measure the fraction of pixels satisfying
\begin{equation}
|\hat d - d| \leq z_p \hat{\sigma},
\end{equation}
where \(z_p\) is the corresponding Gaussian interval coefficient and \(\hat{\sigma}\) is the predicted standard deviation. Perfect calibration would produce empirical coverage close to the nominal confidence level. In our experiments, the empirical coverage curves lie above the ideal diagonal, indicating underconfidence: the predicted uncertainty overestimates the actual error spread.

\begin{table}[ht]
  \caption{Coverage-based expected calibration error for FreqLF+. Lower is better.}
  \label{tab:app_uncertainty_ece}
  \centering
  \small
  \setlength{\tabcolsep}{8pt}
  \renewcommand{\arraystretch}{0.95}
  \begin{tabular}{lc}
    \toprule
    Scene & ECE \\
    \midrule
    Backgammon & 0.480 \\
    Dots       & 0.487 \\
    Pyramids   & 0.496 \\
    Stripes    & 0.482 \\
    \midrule
    Mean       & 0.486 \\
    \bottomrule
  \end{tabular}
\end{table}

\begin{figure}[ht]
  \centering
  \includegraphics[width=\linewidth]{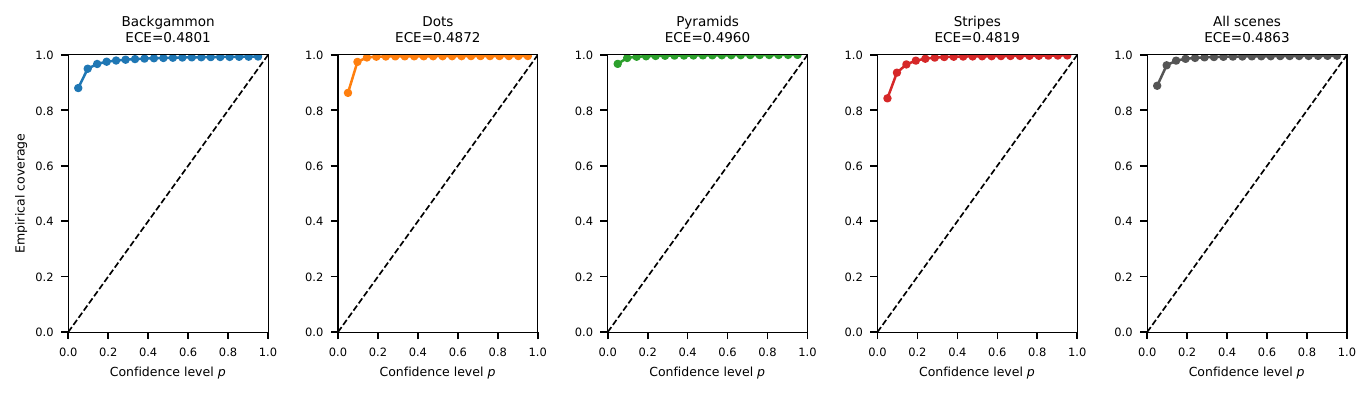}
  \caption{Coverage-based calibration for FreqLF+. The empirical coverage curves lie above the ideal diagonal, indicating underconfidence: the predicted uncertainty overestimates the actual error spread.}
  \label{fig:app_uncertainty_calibration}
\end{figure}

Overall, these results show that the probabilistic decoder provides useful qualitative uncertainty maps, but the predicted variance should not be interpreted as a calibrated confidence estimate. Calibrated uncertainty estimation, including improved uncertainty-error alignment and calibration-error reduction, is left for future work.

\subsection{Additional Qualitative Results}
\label{app:qualitative}

\begin{figure}[ht]
  \centering
  \includegraphics[width=\linewidth]{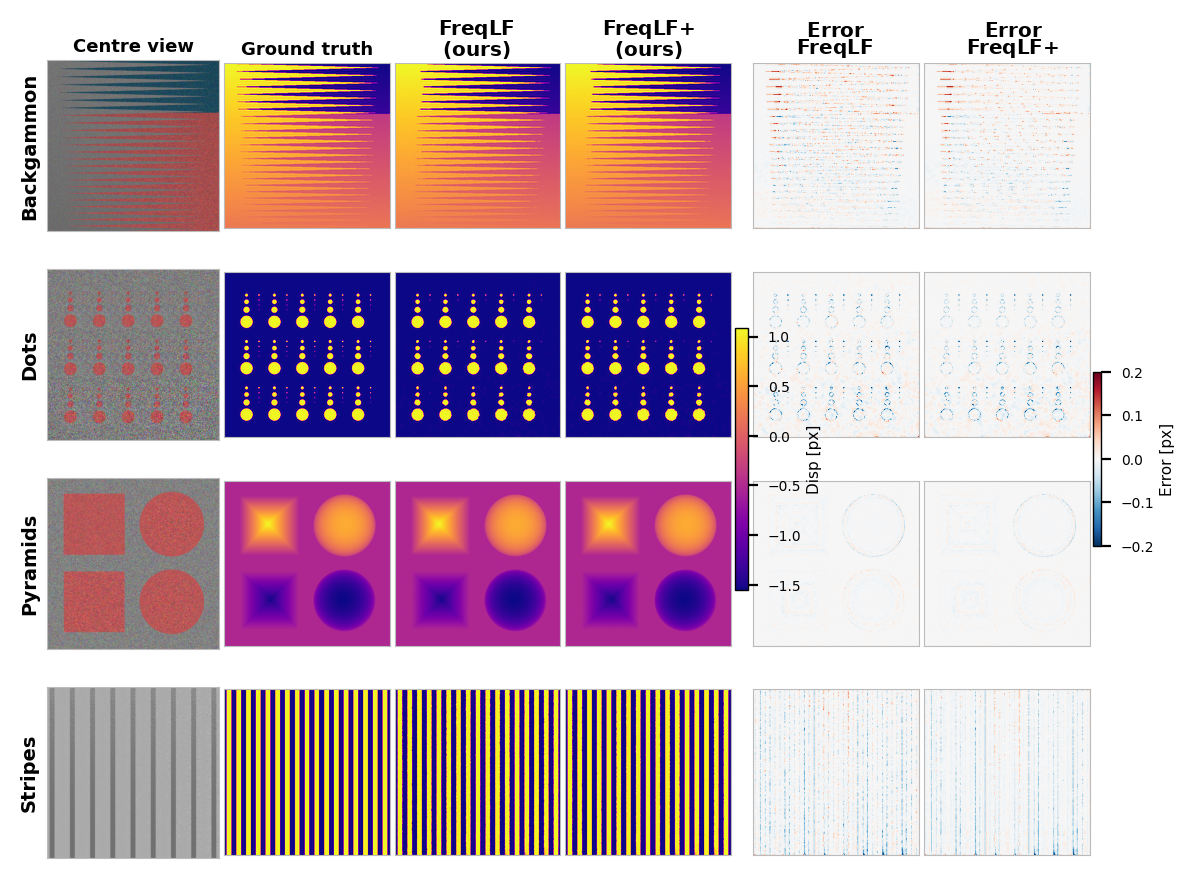}
  \caption{Additional qualitative disparity results.}
  \label{fig:app_qualitative_results}
\end{figure}

Figure~\ref{fig:app_qualitative_results} shows additional qualitative results. The visual examples include the central view, ground-truth disparity, predicted disparity, absolute error, and uncertainty map. These examples complement the quantitative results in the main paper by showing where errors concentrate spatially.

The qualitative results show that most errors occur near disparity discontinuities, repetitive structures, and ambiguous angular evidence. Pyramids is generally easier because the geometry is smooth and texture is regular, while Backgammon, Dots, and Stripes expose different failure modes: matching ambiguity, sparse repeated structures, and high-frequency stripe boundaries.

\subsection{Computational Details and Complexity}
\label{app:computational_details}

\paragraph{Runtime and memory protocol.}
Runtime and memory measurements are performed on a single NVIDIA A100 GPU with 80 GB HBM2e memory. For each input resolution, one warm-up pass is performed before measurement. The reported runtime is the mean over 10 repeated runs. Peak GPU memory is measured during inference. The same trained model is evaluated at different spatial resolutions.

\paragraph{Complexity.}
Let \(N=HW\) denote the number of spatial pixels. The dense encoder and decoder scale approximately linearly with \(N\). In each hybrid operator layer, the FFT-based Fourier branch contributes an \(O(N\log N)\) term, while the local convolutional branch scales linearly in \(N\) for fixed channel count and kernel size. Thus, the dominant spatial scaling of the base FreqLF backbone is approximately \(O(N\log N)\). This scaling is one reason the base model remains tractable at resolutions beyond the native \(512\times512\) benchmark resolution.

\paragraph{Memory behavior.}
The base FreqLF model avoids explicit disparity-volume construction and therefore does not allocate memory proportional to a dense set of candidate disparity hypotheses. Instead, its main memory usage comes from the spatial feature maps, latent field, FFT buffers, and decoder queries. FreqLF+ requires more memory because its enhanced front-end processes richer angular evidence before the same Fourier-local backbone and decoder are applied.

\paragraph{FreqLF+ overhead.}
FreqLF+ improves accuracy by using a stronger occlusion-aware front-end. This front-end processes more angular evidence and substantially increases runtime and memory consumption. In the main paper, FreqLF+ is therefore interpreted as a modular enhanced variant rather than as the primary cost-volume-free model. The base FreqLF model is the variant used to support the cost-volume-free field-learning claim.

\paragraph{Runtime interpretation.}
Reported runtimes for prior methods are taken from their original reports when available and may not be measured on identical hardware. Therefore, they should be interpreted only as indicative values. The controlled runtime and memory analysis for our own models is reported separately in the main paper.

\subsection{Discussion, Scope, and Limitations}
\label{app:discussion_limitations}

\paragraph{Scope of the cost-volume-free claim.}
The base FreqLF model avoids explicit cost-volume construction: it does not enumerate candidate disparities and does not build a multi-view matching volume over discrete disparity hypotheses. Instead, it encodes angular geometry into EPI-derived spatial features and updates the resulting latent field through Fourier-local layers before coordinate-conditioned decoding. The cost-volume-free claim therefore refers to the base FreqLF model. FreqLF+ is included only as an enhanced front-end variant to test whether the proposed Fourier-local backbone benefits from stronger angular features.

\paragraph{Dependence on light-field calibration.}
FreqLF relies on EPI-derived angular structure and therefore assumes that the light-field views are sufficiently calibrated and rectified so that EPI line-slope patterns provide reliable disparity cues. Inaccurate camera calibration, view misalignment, or non-uniform angular sampling can distort EPI geometry and may degrade performance. Cost-volume-based approaches can be more robust in such settings because they explicitly search over candidate disparities and aggregate matching evidence across views. In contrast, the base FreqLF model trades explicit disparity-volume construction for Fourier-local updates over EPI-derived latent fields. Improving robustness to imperfect calibration is an important direction for future work.

\paragraph{Why Fourier-local processing?}
Light-field disparity contains both low-frequency and high-frequency structure. Smooth surfaces, planar regions, and textureless areas require long-range consistency, while occlusion boundaries, thin structures, and foreground-background transitions require local precision. The Fourier branch provides global interaction through retained low-frequency modes, whereas the local CNN branch preserves spatially localized detail. The ablation results support this design: removing the Fourier branch weakens global field modeling, while removing the local branch produces a substantially larger degradation because fine disparity detail is no longer sufficiently preserved.

\paragraph{Relation to Fourier Neural Operators.}
FreqLF is inspired by the operator-learning view of Fourier-domain feature updates, but it is not simply a generic hybrid Fourier-local backbone applied to light-field data. The model is specifically conditioned on light-field geometry through horizontal and vertical EPI stacks. It combines EPI-derived angular features, central-view appearance features, local convolutional refinement, and coordinate-conditioned probabilistic decoding. The goal is therefore not only to apply a Fourier operator, but to use Fourier-local latent field learning as a disparity-specific alternative to explicit disparity-volume construction.

\paragraph{Relation to implicit neural representations.}
FreqLF uses query coordinates in the decoder, but it should not be interpreted as a fully implicit continuous scene representation. The latent representation is computed on an \(H\times W\) spatial grid, and the decoder obtains features by bilinear sampling from this grid. Thus, FreqLF is a coordinate-conditioned decoder over a grid-based latent field. This design allows coordinate injection and probabilistic prediction while retaining efficient dense image-grid processing.

\paragraph{Role of the probabilistic decoder.}
The probabilistic decoder is used to obtain a mixture-based disparity estimate and auxiliary uncertainty maps. The mixture mean is used as the final disparity prediction for all benchmark metrics. In this work, uncertainty is treated as a diagnostic signal rather than as a reliable calibrated confidence estimate. The predicted variance shows weak correlation with absolute error, and coverage-based calibration indicates underconfidence. More rigorous uncertainty modeling and calibration are therefore left for future work.

\paragraph{Limitations and future work.}
FreqLF is primarily evaluated on the HCI 4D Light Field Benchmark, and broader validation on real-world light-field data remains future work. Real-world light-field data may introduce calibration errors, sensor noise, sparse angular sampling, non-Lambertian effects, and domain shifts that are not fully captured by the benchmark. Boundary regions and thin structures remain challenging, as indicated by the region-specific diagnostics, although the Fourier-local design improves performance on high-frequency scenes such as Stripes. FreqLF+ shows that stronger angular encoding can improve accuracy, but it also increases runtime and memory consumption. Future work will focus on boundary-aware refinement, improved robustness to imperfect light-field calibration, improved uncertainty calibration, more efficient enhanced front-ends, and validation on real-world light-field acquisitions.



\end{document}